\documentclass[10pt,twocolumn,letterpaper]{article}

\usepackage{cvpr}
\usepackage{times}
\usepackage{epsfig}
\usepackage{graphicx}
\usepackage{amsmath}
\usepackage{amssymb}

\usepackage{subcaption}
\usepackage{algorithm}
\usepackage{algorithmic}

\usepackage{mathtools}
\usepackage{array}
\usepackage{helvet}
\usepackage{courier}
\usepackage{latexsym}
\usepackage{url}
\usepackage{amsthm}
\usepackage{color}

\usepackage{booktabs}

\usepackage[breaklinks=true,letterpaper=true,bookmarks=false]{hyperref}

\DeclareMathOperator{\E}{\mathbb{E}}
\DeclareMathOperator{\R}{\mathbb{R}}
 
\newtheorem{theorem}{Theorem}

\newtheorem{proposition}{Proposition}
\newtheorem{corollary}{Corollary}

\graphicspath{ {images/}{syntheticExp/} }

\def\tabref#1{Tab.~\ref{#1}}

\cvprfinalcopy

\begin{document} 

\title{Multi-Agent Diverse Generative Adversarial Networks}

\author{
Arnab Ghosh\thanks{Joint first author. This is an updated version of our CVPR'18 paper with the same title. In this version, we also introduce MAD-GAN-Sim in Appendix~\ref{sec:competing}.
} \\
University of Oxford, UK\\
{\tt\small arnabg@robots.ox.ac.uk}
\and
Viveka Kulharia\footnotemark[1]  \\
University of Oxford, UK\\
{\tt\small viveka@robots.ox.ac.uk}
\and
Vinay Namboodiri\\
IIT Kanpur, India \\
{\tt\small vinaypn@iitk.ac.in}
\and
Philip H.S. Torr\\
University of Oxford, UK\\
{\tt\small philip.torr@eng.ox.ac.uk}
\and
Puneet K. Dokania \\
University of Oxford, UK \\
{\tt\small puneet@robots.ox.ac.uk}
}

\maketitle
\begin{abstract}
We propose MAD-GAN, an intuitive generalization to the Generative Adversarial Networks (GANs) and its {\em conditional variants} to address the well known problem of {\em mode collapse}. First, MAD-GAN is a multi-agent GAN architecture incorporating multiple generators and one discriminator. Second, to enforce that different generators capture diverse high probability modes, the discriminator of MAD-GAN is designed such that along with finding the real and fake samples, it is also required to identify the generator that generated the given fake sample. Intuitively, to succeed in this task, the discriminator must learn to push different generators towards different identifiable modes. 
We perform extensive experiments on synthetic and real datasets and compare MAD-GAN with different variants of GAN. 
We show high quality diverse sample generations for challenging tasks such as image-to-image translation and face generation. In addition, we also show that MAD-GAN is able to disentangle different modalities when trained using highly challenging diverse-class dataset (\eg dataset with images of forests, icebergs, and bedrooms). In the end, we show its efficacy on the unsupervised feature representation task. In Appendix, we introduce a similarity based competing objective (MAD-GAN-Sim) which encourages different generators to generate diverse samples based on a user defined similarity metric. We show its performance on the image-to-image translation, and also show its effectiveness on the unsupervised feature representation task.

\end{abstract}
\begin{figure}
    \centering
    \includegraphics[width=\linewidth]{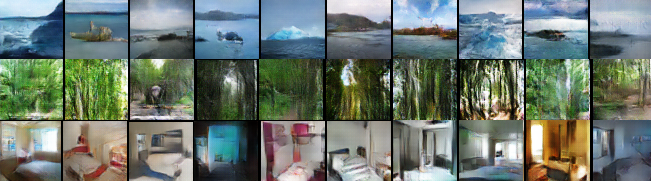}
    \caption{Diverse-class data generation using MAD-GAN. Diverse-class dataset contains images from different classes/modalities (in this case, forests, icebergs, and bedrooms). Each row represents generations by a particular generator and each column represents generations for a given random noise input $z$. As shown, once trained using this dataset, generators of MAD-GAN are able to disentangle different modalities, hence, each generator is able to generate images from a particular modality.
    }\label{fig:madgan_multiview}
    \vspace{-4mm}
\end{figure}

\section{Introduction}

Generative models have attracted considerable attention recently. The underlying idea behind such models is to attempt to capture the distribution of high-dimensional data such as images and texts. Though these models are highly useful in various applications, it is computationally expensive to train them as they require intractable integration in a very high-dimensional space. This drastically limits their applicability. However, recently there has been considerable progress in {\em deep generative models} -- conglomerate of deep neural networks and generative models -- as they do not explicitly require the intractable integration, and can be efficiently trained using back-propagation algorithm. Two such famous examples are Generative Adversarial Networks (GANs)~\cite{goodfellow2014generative} and Variational Autoencoders~\cite{kingma2013auto}.

In this paper we focus on GANs as they are known to produce sharp and plausible images. Briefly, GANs employ a generator and a discriminator where both are involved in a minimax game. The task of the discriminator is to learn the difference between {\em real} samples (from true data distribution $p_d$) and {\em fake} samples (from generator distribution $p_g$). Whereas, the task of the generator is to maximize the mistakes of the discriminator. At convergence, the generator learns to produce real looking images. A few successful applications of GANs are video generation~\cite{vondrick2016generating}, image inpainting~\cite{pathak2016context}, image manipulation~\cite{zhu2016generative}, 3D object generation~\cite{wu2016learning}, interactive image generation using few brush strokes~\cite{zhu2016generative}, image super-resolution~\cite{ledig2016photo}, diagrammatic abstract reasoning~\cite{ghosh2017contextual} and conditional GANs~\cite{mirza2014conditional,reed2016generative}.

Despite the remarkable success of GAN, it suffers from the major problem of {\em mode collapse}~\cite{arjovsky2016towards,che2016mode,chen2016infogan,metz2017unrolledGAN,salimans2016improved}. Though, theoretically, convergence guarantees the generator learning the true data distribution. However, practically, reaching the true equilibrium is difficult and not guaranteed, which potentially leads to the aforementioned problem of mode collapse. Broadly speaking, there are two schools of thought to address the issue: (1) improving the learning of GANs to reach better optima~\cite{arjovsky2016towards,metz2017unrolledGAN,salimans2016improved}; and (2) explicitly enforcing GANs to capture diverse modes~\cite{che2016mode,chen2016infogan,liu2016coupled}. Here we focus on the latter.

\begin{figure}[ht]
    \centering
     \includegraphics[scale=0.38]{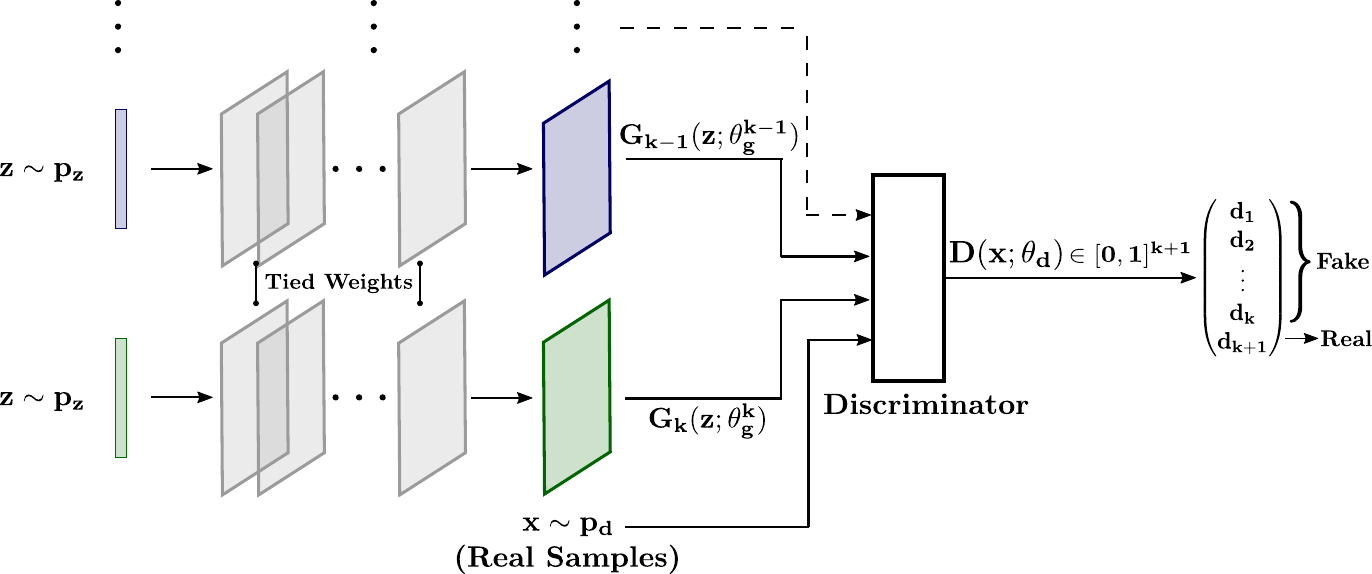}
	\caption{Multi-Agent Diverse GAN (MAD-GAN). The discriminator outputs $k+1$ softmax scores signifying the probability of its input sample being from either one of the $k$ generators or the real distribution. 
	}
   \label{fig:multiAgentGAN}
   \vspace{-4mm}
\end{figure}

Borrowing from the multi-agent algorithm~\cite{abadi2016learning} and coupled GAN~\cite{liu2016coupled}, we propose to use multiple generators with one discriminator. We call this framework the Multi-Agent GAN architecture, as shown in Fig.~\ref{fig:multiAgentGAN}. In detail, similar to the standard GAN, the objective of each generator here is to maximize the mistakes of the {\em common} discriminator. Depending on the task, it might be useful for different generators to share information. This is done using the initial layer parameters of generators. Another reason behind sharing these parameters is the fact that initial layers capture low-frequency structures which are almost the same for a particular type of dataset (for example, faces), therefore, sharing them reduces redundant computations. However, when the dataset contains images from completely different modalities, one can avoid sharing these parameters. Naively using multiple generators may lead to the {\em trivial solution} where all the generators learn to generate {\em similar} samples. To resolve this issue and generate different visually plausible samples capturing diverse high probability modes, we propose to modify the objective function of the discriminator. In the modified objective, along with finding the real and the fake samples, the discriminator also has to correctly identify the generator that generated the given fake sample. Intuitively, in order to succeed in this task, the discriminator must learn to push generations corresponding to different generators towards different identifiable modes. Combining the Multi-Agent GAN architecture with the diversity enforcing term allows us to generate diverse plausible samples, thus the name Multi-Agent Diverse GAN (MAD-GAN).

As an example, an intuitive setting where mode collapse occurs is when a GAN is trained on a dataset containing images from different modalities/classes. For example, a diverse-class dataset containing images such as forests, iceberg, and bedrooms. This is of particular interest as it not only requires the model to disentangle intra-class variations, it also requires inter-class disentanglement. Fig.~\ref{fig:madgan_multiview} demonstrates the surprising effectiveness of MAD-GAN in this challenging setting. Generators among themselves are able to disentangle inter-class variations, and each generator is also able to capture intra-class variations.

In addition, we analyze MAD-GAN through extensive experiments and compare it with several variants of GAN. First, for the proof of concept, we perform experiments in controlled settings using synthetic dataset (mixture of Gaussians), and complicated Stacked/Compositional MNIST datasets with hand engineered modes. In these settings, we empirically show that our approach outperforms all other GAN variants we compare with, and is able to generate high quality samples while capturing large number of modes. In a more realistic setting, we show high quality diverse sample generations for the challenging tasks of {\em image-to-image translation}~\cite{isola2016image2image} (conditional GAN) and {\em face generation}~\cite{chen2016infogan,radford2015unsupervised}. Using the SVHN dataset~\cite{Netzer11SVHN}, we also show the efficacy of our framework for learning the feature representation in an unsupervised setting.

We also provide theoretical analysis of this approach and show that the proposed modification in the objective of discriminator allows generators to learn together as a mixture model where each generator represents a mixture component. We show that at convergence, the global optimum value of $-(k+1)\log(k+1) + k \log k$ is achieved, where $k$ is the number of generators. 

\section{Related Work}
The recent work called InfoGAN~\cite{chen2016infogan} proposed an information-theoretic extension to GANs in order to address the problem of mode collapse. Briefly, InfoGAN disentangles the latent representation by assuming a factored representation of the latent variables. In order to enforce that the generator learns factor specific generations, InfoGAN maximizes the mutual information between the factored latents and the generator distribution. 
Che \etal~\cite{che2016mode} proposed a mode regularized GAN (ModeGAN) which uses an encoder-decoder paradigm. The basic idea behind ModeGAN is that if a sample from the true data distribution $p_d$ belongs to a particular mode, then the sample generated by the generator (fake sample) when the true sample is passed through the encoder-decoder is likely to belong to the same mode. ModeGAN assumes that there exists enough true samples from a mode for the generator to be able to capture it.
Another work by Metz \etal~\cite{metz2017unrolledGAN} proposed a surrogate objective for the update of the generator with respect to the unrolled optimization of the discriminator (UnrolledGAN) to address the issue of convergence of the training process of GANs. This improves the training process of the generator which in turn allow the generators to explore better coverage to true data distribution. 

Liu \etal~\cite{liu2016coupled} presented Coupled GAN, a method for training two generators with shared parameters to learn the joint distribution of the data. The shared parameters guide both the generators towards similar subspaces but since they are trained independently on two domains, they promote diverse generations. Durugkar \etal~\cite{durugkar2016generative} proposed a model with multiple discriminators whereby an ensemble of multiple discriminators have been shown to stabilize the training of the generator by guiding it to produce better samples. 

W-GAN~\cite{arjovsky2017wasserstein} is a recent technique which employs integral probability metrics based on the earth mover distance rather than the JS-divergences that the original GAN uses. BEGAN~\cite{berthelot2017began} builds upon W-GAN using an autoencoder based equilibrium enforcing technique alongside the Wasserstein distance. DCGAN~\cite{radford2015unsupervised} was a seminal technique which used a fully convolutional generator and discriminator for the first time along with the introduction of batch normalization thus stabilizing the training procedure, and was able to generate compelling generations. GoGAN~\cite{juefei2017gang} introduced a training procedure for the training of the discriminator using a maximum margin formulation alongside the earth mover distance based on the Wasserstein-1 metric.~\cite{arora2017generalization} introduced a technique and theoretical formulation stating the importance of multiple generators and discriminators in order to completely model the data distribution. In terms of employing multiple generators, our work is closest to~\cite{arora2017generalization,liu2016coupled,ghosh2016message}. However, while using multiple generators, our method explicitly enforces them to capture diverse modes. 

\section{Preliminaries}
Here we present a brief review of GANs~\cite{goodfellow2014generative}. Given a set of samples $\mathcal{D} = (x_i)_{i=1}^n$ from the true data distribution $p_d$, the GAN learning problem is to obtain the optimal parameters $\theta_g$ of a generator $G(z;\theta_g)$ that can sample from an approximate data distribution $p_g$, where $z \sim p_z$ is the prior input noise (\eg samples from a normal distribution). In order to learn the optimal $\theta_g$, the GAN objective (Eq.~\eqref{eq:ganObjective}) employs a discriminator $D(x; \theta_d)$ that learns to differentiate between a {\em real} (from $p_d$) and a {\em fake} (from $p_g$) sample $x$. The overall GAN objective is:
\begin{align}
    \label{eq:ganObjective}
    \min_{\theta_g}\max_{\theta_d}\; & V(\theta_d, \theta_g) := \E_{x \sim p_{d}} \log D(x; \theta_d) \nonumber \\&
    + \E_{z \sim p_z } \log \big( 1 - D(G(z; \theta_g) ; \theta_d)\big) 
\end{align}
The above objective is optimized in a block-wise manner where $\theta_d$ and $\theta_g$ are optimized one at a time while fixing the other. For a given sample $x$ (either from $p_d$ or $p_g$) and the parameter $\theta_d$, the function $D(x; \theta_d) \in [0, 1]$ produces a score that represents the probability of $x$ belonging to the true data distribution $p_d$ (or probability of it being real). The objective of the discriminator is to learn parameters $\theta_d$ that maximizes this score for the true samples (from $p_d$) while minimizing it for the fake ones $\tilde{x} = D(z; \theta_g)$ (from $p_g$). In the case of generator, the objective is to minimize $\E_{z \sim p_z}\log \big( 1 - D(G(z; \theta_g) ; \theta_d) \big)$, equivalently maximize $\E_{z \sim p_z} \log D(G(z; \theta_g) ; \theta_d)$. Thus, the generator learns to maximize the scores for the fake samples (from $p_g$), which is exactly the opposite to what discriminator is trying to achieve. In this manner, the generator and the discriminator are involved in a minimax game where the task of the generator is to maximize the mistakes of the discriminator. Theoretically, at equilibrium, the generator learns to generate real samples, which means $p_g = p_d$.

\section{Multi-Agent Diverse GAN}
In the GAN objective, one can argue that the task of a generator is much harder than that of the discriminator as it has to produce real looking images to maximize the mistakes of the discriminator. This, along with the minimax nature of the objective raise several challenges for GANs~\cite{arjovsky2016towards,che2016mode,chen2016infogan,metz2017unrolledGAN,salimans2016improved}: (1) mode collapse; (2) difficult optimization; and (3) trivial solution. In this work we propose a new framework to address the first challenge of {\em mode collapse} by increasing the capacity of the generator while using well known tricks to partially avoid other challenges~\cite{arjovsky2016towards}. 

Briefly, we propose a {\em Multi-Agent GAN architecture} that employs multiple generators and one discriminator in order to generate different samples from high probability regions of the true data distribution. 
In addition, theoretically, we show that our formulation allows generators to act as a mixture model with each generator capturing one component.

\subsection{Multi-Agent GAN Architecture}
\label{subsection:architecture}
Here we describe our proposed architecture (Fig.~\ref{fig:multiAgentGAN}). It involves $k$ generators and one discriminator. In the case of homogeneous data (all the images belong to same class, \eg faces or birds), we allow all the generators to share information by tying most of the initial layer parameters. This is essential to avoid redundant computations as initial layers of a generator capture low-frequency structures which are almost the same for a particular type of dataset. This also allows different generators to converge faster. However, in the case of {\em diverse-class} data (\eg dataset with a mixture of different classes such as forests, icebergs \etc), it is necessary to avoid sharing these parameters to allow each generator to capture content specific structures. Thus, the extent to which one should share these parameters depends on the task at hand. 

More specifically, given $z \sim p_z$ for the $i$-th generator, similar to the standard GAN, the first step involves generating a sample (for example, an image) $\tilde{x}_i$. Since each generator receives the same latent input sampled from the same distribution, naively using this simple approach may lead to the {\em trivial solution} where all the generators learn to generate similar samples. In what follows, we propose an intuitive solution to avoid this issue and allow the generators to capture diverse modes.

\subsection{Enforcing Diverse Modes}
\label{sec:diverseModes}
\begin{figure}
    \centering
     \includegraphics[width=0.5\linewidth]{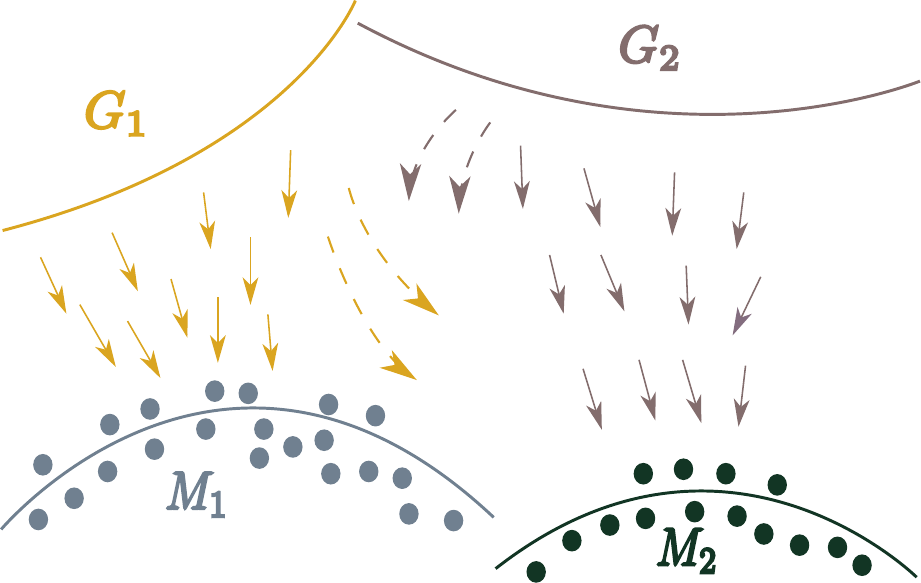}
	\caption{Visualization of different generators getting pushed towards different modes. Here, $M_1$ and $M_2$ could be a cluster of modes where each cluster itself contains many modes. The arrows abstractly represent generator specific gradients for the purpose of building intuition.}
   \label{fig:modeVisualization}
   \vspace{-4mm}
\end{figure}

Inspired by the discriminator formulation for the semi-supervised learning~\cite{salimans2016improved}, we use a generator identification based objective function that, along with minimizing the score $D(\tilde{x}; \theta_d)$, requires the discriminator to identify the generator that generated the given fake sample $\tilde{x}$. In order to do so, as opposed to the standard GAN objective function where the discriminator outputs a scalar value, we modify it to output $k+1$ soft-max scores. In more detail, given the set of $k$ generators, the discriminator produces a soft-max probability distribution over $k+1$ classes.  The score at $(k+1)$-th index $(D_{k+1}(.))$ represents the probability that the sample belongs to the true data distribution and the score at $j \in \{1, \dots, k\}$-th index represents the probability of it being generated by the $j$-th generator. Under this setting, while learning $\theta_d$, we optimize the cross-entropy between the soft-max output of the discriminator and the Dirac delta distribution $\delta \in \{0,1\}^{k+1}$, where for $j \in \{1, \dots, k\}$, $\delta(j) = 1$ if the sample belongs to the $j$-th generator, otherwise $\delta(k+1) = 1$. Thus, the objective of the discriminator, which is optimizing $\theta_d$ while keeping $\theta_g$ constant (refer Eq.~(\ref{eq:ganObjective})), is modified to:
\begin{align*}
    \max_{\theta_d} \E_{x \sim p} H(\delta, D(x; \theta_d))
\end{align*}
where, $Supp(p) = \cup_{i=1}^k Supp(p_{g_i}) \cup Supp(p_d) $ and $H(.,.)$ is the negative of the cross entropy function. Intuitively, in order to correctly identify the generator that produced a given fake sample, the discriminator must learn to push different generators towards different identifiable modes. However, the objective of each generator remains the same as in the standard GAN. Thus, for the $i$-th generator, the objective is to minimize the following:
\begin{align}
\mathbb{E}_{x\sim p_{d}}\log D_{k+1}(x; \theta_d) \hspace{-1mm}+ \hspace{-1mm}\mathbb{E}_{z \sim p_z}\log (1- D_{k+1}(G_i(z; \theta_g^i); \theta_d)) \nonumber
\end{align}
To update the parameters, the gradient for each generator is simply computed as $\nabla_{\theta_g^i} \log (1 - D_{k+1}(G_i(z; \theta_g^i); \theta_d))$. Notice that all the generators in this case can be updated in parallel. For the discriminator, given $x \sim p$ (can be real or fake) and corresponding $\delta$, the gradient is $\nabla_{\theta_d}  \log D_j(x; \theta_d)$, where $D_j(x; \theta_d)$ is the $j$-th index of  $D(x; \theta_d)$ for which $\delta(j) = 1$. Therefore, using this approach requires {\em very minor modifications to the standard GAN optimization algorithm} and can be easily used with different variants of GAN. An intuitive visualization is shown in Fig.~\ref{fig:modeVisualization}.

Theorem~\ref{theo:optimalConditionGenId} shows that the above objective function actually allows generators to form a mixture model where each generator represents a mixture component and the global optimum of $-(k+1) \log (k+1) + k \log k$ is achieved when $p_d = \frac{1}{k} \sum_{i=1}^k p_{g_i}$. Notice that, at $k=1$, which is the case with one generator, we obtain exactly the same Jensen-Shannon divergence based objective function as shown in~\cite{goodfellow2014generative} with the optimal value of $-\log 4$.

\begin{theorem}
\label{theo:optimalConditionGenId}
Given the optimal discriminator, the objective for training the generators boils down to minimizing
\begin{align}
\label{eq:klDivProof}
& KL \Big(p_d(x) ||  p_{avg}(x) \Big) + k KL \Big( \frac{1}{k} \sum_{i=1}^k p_{g_i} (x) ||  p_{avg}(x) \Big)  \nonumber \\
&-(k+1) \log (k+1) + k \log k
\end{align}
where, $p_{avg} (x) = \frac{p_d(x) + \sum_{i=1}^k p_{g_i}(x)}{k+1}$. The above objective function obtains its global minimum if $p_d = \frac{1}{k} \sum_{i=1}^k p_{g_i}$ with the objective value of $-(k+1) \log (k+1) + k \log k$.
\begin{proof}
The joint objective of all the generators is to minimize the following:
\begin{align}
\mathbb{E}_{x\sim p_{d}}\log D_{k+1}(x) + \sum_{i=1}^{k}\mathbb{E}_{x \sim p_{g_i}}\log (1- D_{k+1}(x)) \nonumber 
\end{align}
Using Corollary~\ref{cor:optimalDiscrminator}, we substitute the optimal discriminator in the above equation and obtain:
\begin{align}
\label{eq:klDivFinal}
& \mathbb{E}_{x\sim p_{d}}\log \Bigg [ \frac{p_d(x)}{p_d(x) + \sum_{i=1}^k p_{g_i}(x)} \Bigg]  + \nonumber \\
&\sum_{i=1}^{k} \mathbb{E}_{x\sim p_{g_i}} \log \Bigg[ \frac{\sum_{i=1}^k p_{g_i}(x)}{p_d(x) + \sum_{i=1}^k p_{g_i}(x)} \Bigg] \nonumber \\
&= \mathbb{E}_{x\sim p_{d}} \log \Bigg [ \frac{p_d(x)}{p_{avg}(x)} \Bigg]  + k \mathbb{E}_{x \sim p_g} \log \Bigg[ \frac{p_g(x)}{p_{avg}(x)} \Bigg] \nonumber \\
&- (k+1) \log (k+1) + k \log k
\end{align}
where, $p_g=\frac{\sum_{i=1}^k p_{g_i}}{k}$ and $p_{avg} (x) = \frac{p_d(x) + \sum_{i=1}^k p_{g_i}(x)}{k+1}$. Note that, Eq.~(\ref{eq:klDivFinal}) is exactly the same as Eq.~(\ref{eq:klDivProof}). When $p_d = \frac{\sum_{i=1}^k p_{g_i}}{k}$, both the KL terms become zero and the global minimum is achieved. 
\end{proof}
\end{theorem}

\begin{corollary}
\label{cor:optimalDiscrminator}
For fixed generators, the optimal distribution learned by the discriminator D has the following form:
\begin{align*}
D_{k+1}(x) &=  \frac{p_d(x)}{p_d(x) + \sum_{i=1}^k p_{g_i}(x)}, \\ \nonumber
D_i(x) &=  \frac{p_{g_i}(x)}{p_d(x) + \sum_{i=1}^k p_{g_i}(x)}, \forall i \in \{1, \cdots, k\}.
\end{align*}
where, $D_i(x)$ represents the $i$-th index of $D(x; \theta_d)$, $p_d$ the true data distribution, and $p_{g_i}$ the distribution learned by the $i$-th generator.
\begin{proof}
For fixed generators, the objective function of the discriminator is to maximize
\begin{align*}
\mathbb{E}_{x\sim p_{d}}\log D_{k+1}(x) + \sum_{i=1}^{k}\mathbb{E}_{x_i\sim p_{g_i}}\log D_i (x_i) 
\end{align*}
where, $\sum_{i=1}^{k+1} D_i(x) = 1$ and $D_i(x) \in [0, 1], \forall i$. The above equation can be written as:
\begin{align}
    \label{eq:optDis2}
& \int_x p_d(x)\log D_{k+1} (x) dx + \sum_{i=1}^k\int_x p_{g_i}(x)\log D_i(x) dx \nonumber \\
& = \int_{x \in p} \sum_{i=1}^{k+1} p_i(x) \log D_i(x) dx
\end{align}
where, $p_{k+1}(x) := p_d(x)$, $p_i(x) := p_{g_i} (x), \forall i \in \{1, \cdots, k\}$, and $Supp(p) = \bigcup_{i=1}^k Supp(p_{g_i}) \bigcup Supp(p_d)$,   Therefore, for a given $x$, the optimum of objective function defined in Eq.~(\ref{eq:optDis2}) with constraints defined above can be obtained using Proposition~\ref{prop:lpCons}.
\end{proof}
\end{corollary}

\begin{proposition}
\label{prop:lpCons}
Given ${\bf y} = (y_1, \cdots, y_n)$, $y_i \geq 0$, and $a_i \in \mathbb{R}$, the optimal solution for the objective function defined below is achieved at $y_i^* = \frac{a_i}{\sum_{i = 1}^n a_i}, \forall i$
\begin{align*}
    \max_{\bf y} \sum_{i=1}^n a_i\log y_i,  \;\; s.t. \; \sum_i^n y_i = 1
\end{align*}
\begin{proof}
    The Lagrangian of the above problem is:
    \begin{align*}
        L({\bf y}, \lambda) = \sum_{i=1}^n a_i\log y_i + \lambda (\sum_{i=1}^n y_i - 1) 
    \end{align*}
    Differentiating w.r.t $y_i$ and $\lambda$, and equating to zero,
    \begin{align*}
        \frac{a_i}{y_i} + \lambda = 0 \; , \; \sum_{i=1}^n y_i - 1 = 0
    \end{align*}
    Solving the above two equations, we obtain $y_i^* = \frac{a_i}{\sum_{i = 1}^n a_i}$.
\end{proof}
\end{proposition}

\newcommand{\addSubFig}[3]{\begin{subfigure}[t]{.33\linewidth}
   \includegraphics[width=\linewidth]{#1}
   \caption{#2}\label{#3}\end{subfigure}
}

%% 1D experiments
\newcommand{\addSubFigEighth}[3]{\begin{subfigure}[t]{.24\linewidth}
   \includegraphics[width=\linewidth]{#1}
   \caption{#2}\label{#3}\vspace{-1mm}\end{subfigure}
}

\newcommand{\addSubFigNineth}[3]{\begin{subfigure}[t]{.19\linewidth}
   \includegraphics[width=\linewidth]{#1}
   \caption{#2}\label{#3}\vspace{-1mm}\end{subfigure}
}

%% 1D experiments
\begin{figure*}
    \centering
    \addSubFigNineth{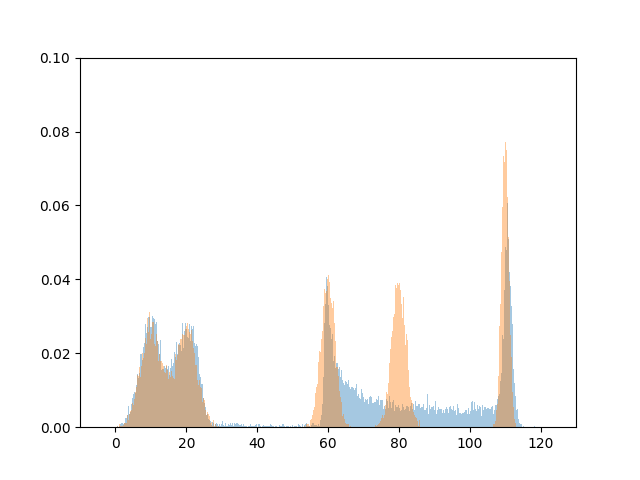}{DCGAN}{fig:DCGAN_1D} 
    \addSubFigNineth{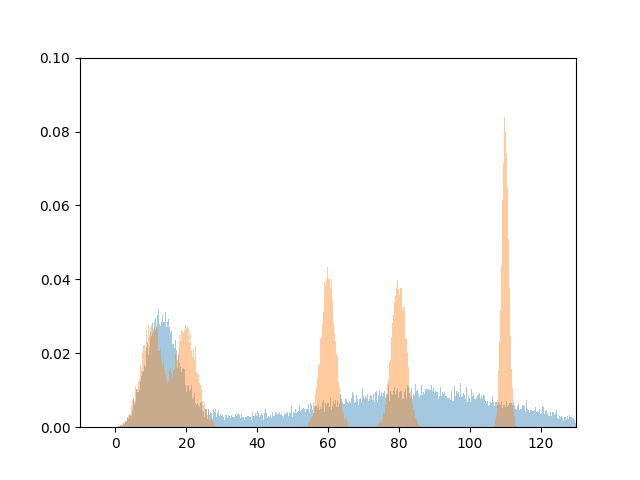}{WGAN}{fig:WGAN_1D} 
    \addSubFigNineth{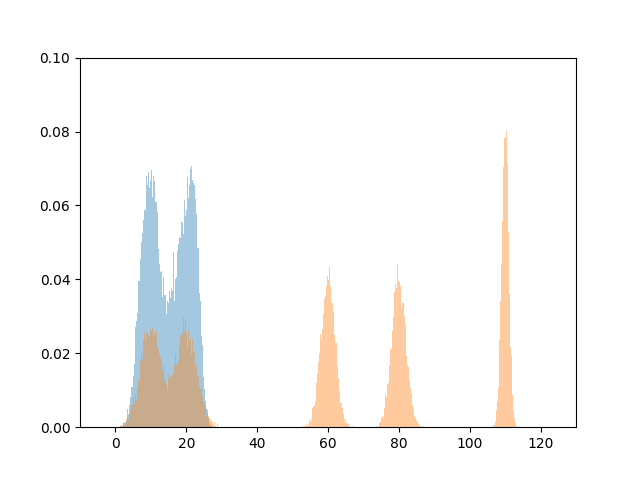}{BEGAN}{fig:BEGAN_1D} 
    \addSubFigNineth{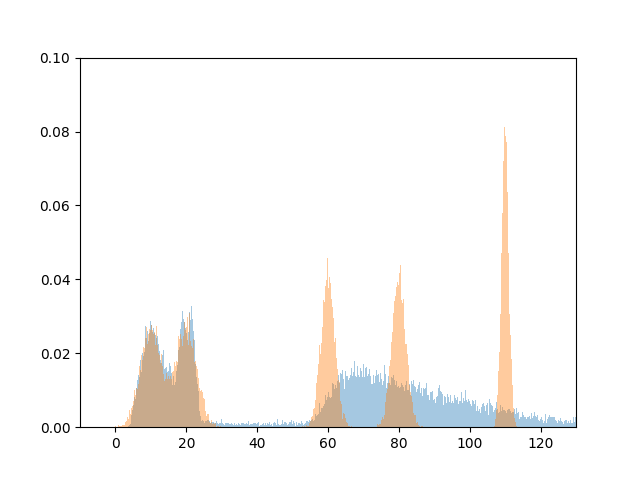}{GoGAN}{fig:GoGAN_1D}\\ 
    \addSubFigNineth{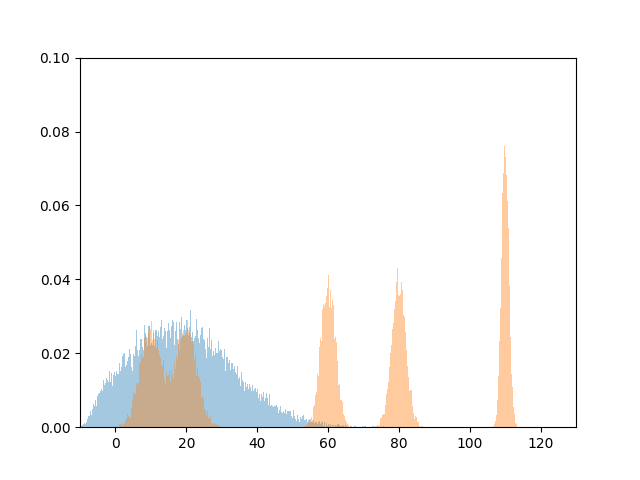}{Unrolled GAN}{fig:Unrolled_1D} 
    \addSubFigNineth{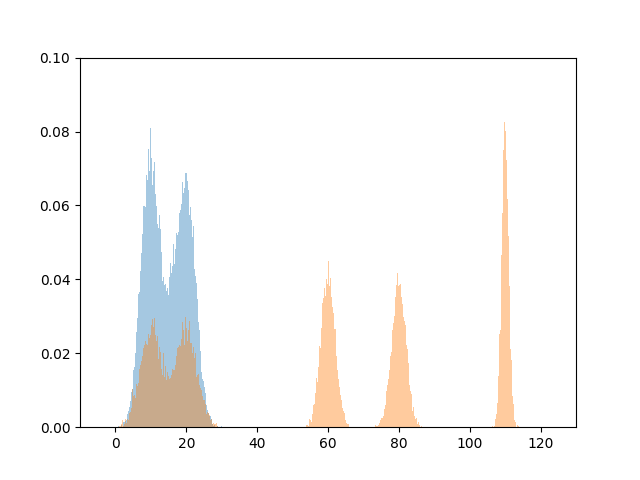}{Mode-Reg DCGAN}{fig:ModeReg_1D} 
    \addSubFigNineth{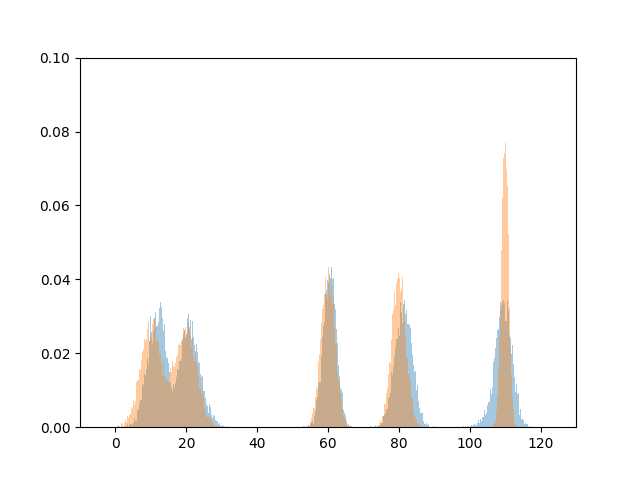}{InfoGAN}{fig:InfoGAN_1D} 
	\addSubFigNineth{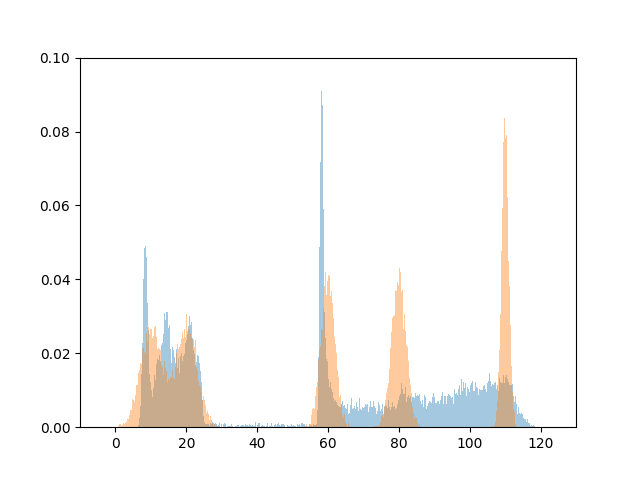}{MA-GAN}{fig:MA-GAN_1D} 
    \addSubFigNineth{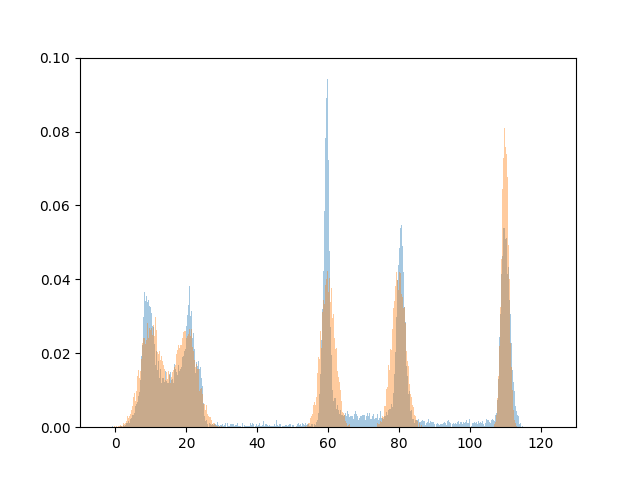}{MAD-GAN (Our)}{fig:MAD-GAN_1D} 
    \caption{A toy example to understand the behaviour of different GAN variants in order to compare with MAD-GAN (each method was trained for 198000 iterations).  The orange bars show the density estimate of the training data and the blue ones for the generated data points. After careful cross-validation, we chose the bin size of $0.1$. }
    \label{fig:toy_1D}
    \vspace{-3mm}
\end{figure*}

\begin{figure*}
    \centering
    \addSubFigEighth{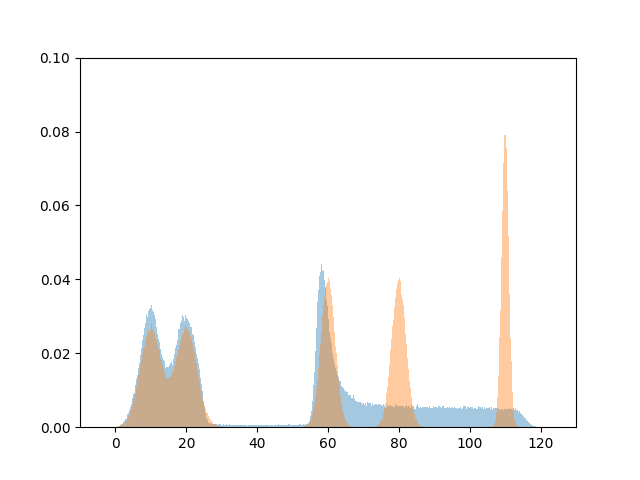}{1 Generator}{fig:MADGAN_1} 
    \addSubFigEighth{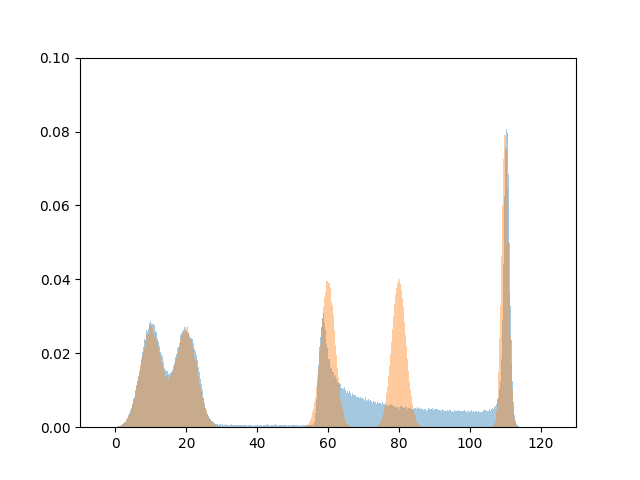}{2 Generators}{fig:MADGAN_2} 
    \addSubFigEighth{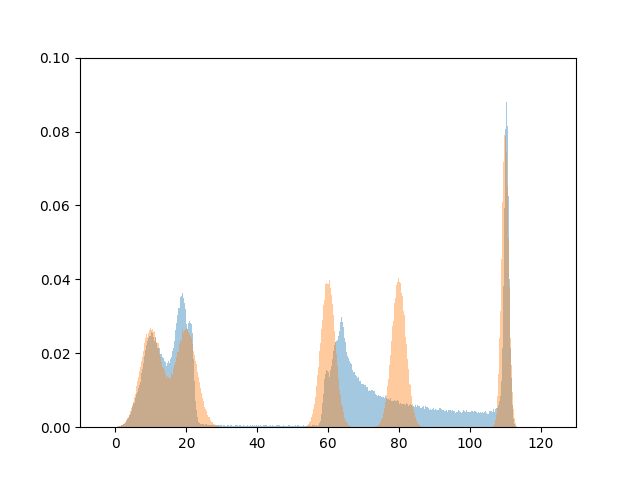}{3 Generators}{fig:MADGAN_3} 
    \addSubFigEighth{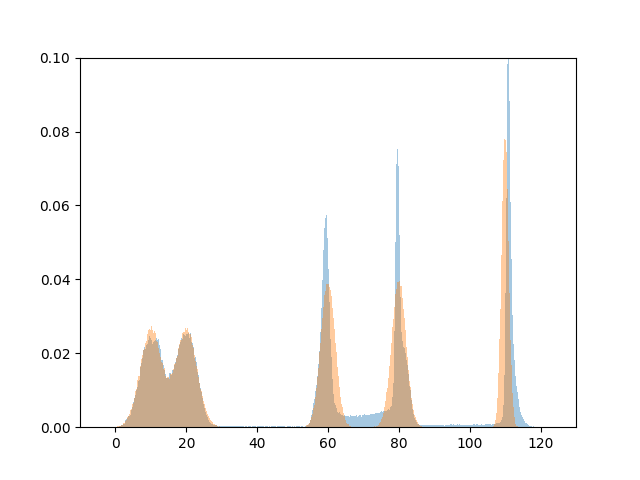}{4 Generators}{fig:MADGAN_4}\\ 
    \addSubFigEighth{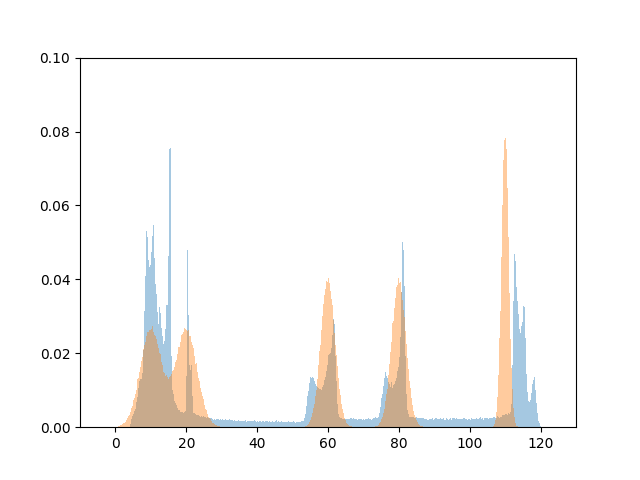}{5 Generators}{fig:MADGAN_5} 
    \addSubFigEighth{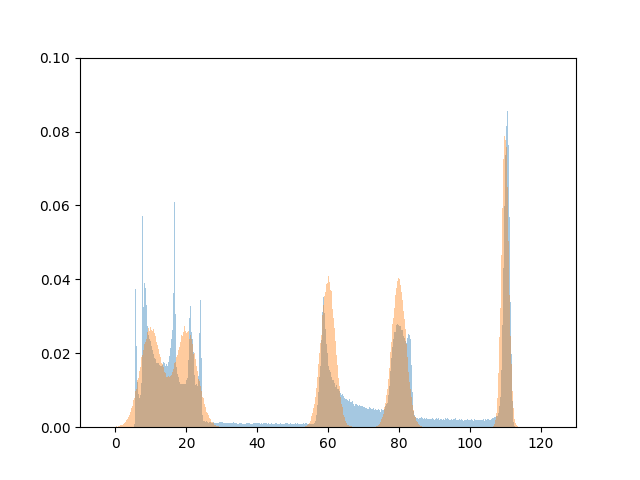}{6 Generators}{fig:MADGAN_6} 
    \addSubFigEighth{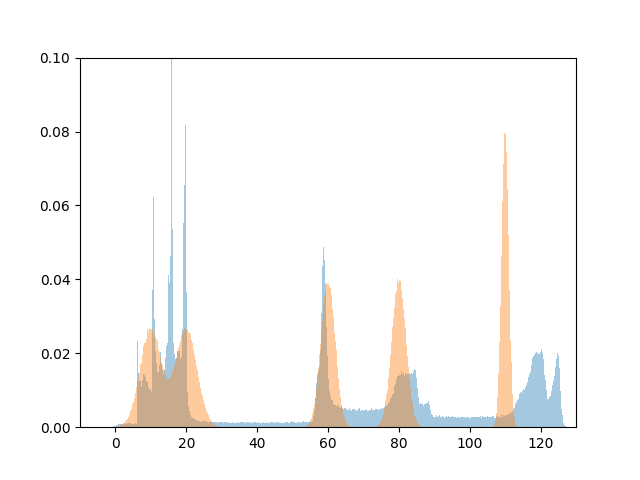}{7 Generators}{fig:MADGAN_7} 
    \addSubFigEighth{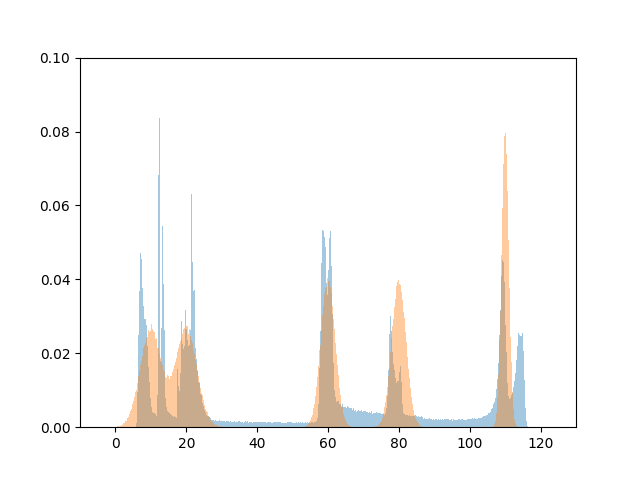}{8 Generators}{fig:MADGAN_8} 
    \caption{A toy example to understand the behavior of MAD-GAN with different number of generators (each method was trained for $1,98,000$ iterations).  The orange bars show the density estimate of the training data and the blue ones for the generated data points. After careful cross-validation, we chose the bin size of $0.1$. }
    \label{fig:toy_1D_multi}
	\vspace{-3mm}
\end{figure*}

\section{Experiments}
We present an extensive quantitative and qualitative analysis of MAD-GAN on various synthetic and real-world datasets. First, we use a simple 1D mixture of Gaussians and also Stacked/Compositional MNIST dataset (1000 modes) to compare MAD-GAN with several known variants of GANs, such as DCGAN \cite{radford2015unsupervised}, WGAN \cite{arjovsky2017wasserstein}, BEGAN \cite{berthelot2017began}, GoGAN \cite{juefei2017gang}, Unrolled GAN \cite{metz2017unrolledGAN}, Mode-Reg GAN \cite{che2016mode} and InfoGAN \cite{chen2016infogan}. Furthermore, we created another baseline, called {\em MA-GAN (Multi-Agent GAN)}, which is a trivial extension of GAN with multiple generators and one discriminator. As opposed to MAD-GAN, MA-GAN has a simple Multi-Agent architecture without modifications to the objective of the discriminator. This comparison allows us to understand the effect of explicitly enforcing diversity in the objective of the MAD-GAN. We use KL-divergence \cite{kullback1951information} and number of modes recovered \cite{che2016mode} as the criterion for comparisons and show superior results compared to all the other methods. Additionally, we show diverse generations for the challenging tasks of {\em image-to-image translation}~\cite{isola2016image2image}, {\em diverse-class data generation}, and face generation. It is non-trivial to devise a metric to evaluate diversity on these high quality generation tasks, so we perform qualitative assessment. Note that, the image-to-image translation objective is known to learn the delta distribution, thus, it is agnostic to the input noise vector. However, we show that MAD-GAN is able to produce highly plausible diverse generations for this task. In the end, we show the efficacy of MAD-GAN in unsupervised feature representation learning task. We provide detailed overview of the architectures, datasets, and the parameters used in our experiments in the Appendix~\ref{sec:architectures}.

In the case of InfoGAN~\cite{chen2016infogan}, we varied the dimension of the categorical variable, depicting the number of modes, to obtain the best cross-validated results.

	\begin{table}
		\centering
		\begin{tabular}{c@{\hspace{2pt}}c@{\hspace{4pt}}c}
		\toprule
			\textbf{GAN Variants} & \textbf{Chi-square}${(\times10^5)}$ & \textbf{KL-Div} \\
			\midrule
			DCGAN~\cite{radford2015unsupervised} & 0.90  & 0.322\\ 
			WGAN~\cite{arjovsky2017wasserstein} & 1.32 & 0.614\\ 
			BEGAN~\cite{berthelot2017began} & 1.06 & 0.944\\ 
			GoGAN~\cite{juefei2017gang} & 2.52 & 0.652\\
			Unrolled GAN~\cite{metz2017unrolledGAN} & 3.98 & 1.321 \\ 
			Mode-Reg DCGAN~\cite{che2016mode}  & 1.02 & 0.927 \\ 			
			InfoGAN~\cite{chen2016infogan}  & 0.83 & 0.21 \\ 
			MA-GAN & 1.39 & 0.526 \\
			\textbf{MAD-GAN (Our)} &  \textbf{0.24} & \textbf{0.145}\\ 
			\bottomrule
		\end{tabular}
		\caption{\label{tab:toy_1D}Synthetic experiment on 1D GMM (Fig.~\ref{fig:toy_1D}).}
		\vspace{-1mm}
	\end{table}

	\begin{table}
		\centering
		\begin{tabular}{c@{\hspace{8pt}}c@{\hspace{8pt}}c} 
			\toprule
			\textbf{\# Generators} & \textbf{Chi-square} ${(\times10^7)}$ & \textbf{KL-Div} \\
			\midrule
			1 & 1.27  & 0.57\\ 
			2 & 1.38 & 0.42\\ 
			3 & 3.15 & 0.71\\
			4 & 0.39 & 0.28\\
			5 & 3.05 & 0.88\\
			6 & 0.54 & 0.29\\
			7 & 0.97 & 0.78\\
			8 & 4.83 & 0.68\\
			\bottomrule
		\end{tabular}
		\caption{\label{tab:toy_1D_multi}Synthetic experiment with different number of MAD-GAN generators (same setup as in Fig.~\ref{fig:toy_1D}).}
		\vspace{-3mm}
	\end{table}

\subsection{Non-Parametric Density Estimation}
\label{sec:toyExperiment}
In order to understand the behavior of MAD-GAN and different state-of-the-art GAN models, we first perform a very simple synthetic experiment, much easier than generating high-dimensional complex images. We consider a distribution of 1D GMM~\cite{bishop2007pattern} having five mixture components with modes at 10, 20, 60, 80 and 110, and standard deviations of 3, 3, 2, 2 and 1, respectively. While the first two modes overlap significantly, the fifth mode stands isolated as shown in Fig.~\ref{fig:toy_1D}. We train different GAN models using $200,000$ samples from this distribution and generate $65,536$ data points from each model. In order to compare the learned distribution with the ground truth distributions, we first estimate them using bins over the data points and create the histograms. These histograms are carefully created using different bin sizes and the best bin (found to be $0.1$) is chosen. Then, we use Chi-square distance and the KL-divergence to compute distance between the two histograms. From Fig.~\ref{fig:toy_1D} and Tab.~\ref{tab:toy_1D} it is evident that MAD-GAN is able to capture all the clustered modes which includes significantly overlapped modes as well. MAD-GAN obtains the minimum value in terms of both Chi-square distance and the KL-divergence. In this experiment, both MAD-GAN and MA-GAN used four generators. In the case of InfoGAN, we used $5$ dimensional categorical variable, which provides the best result. 

To understand the effect of varying the number of generators in MAD-GAN, we use the same synthetic experiment setup, i.e. the real data distribution is same GMM with $5$ Gaussians. For better non-parametric estimation, we use $1$ million sample points from real distribution (instead of $65,536$). We generate equal number of points from each of the generators such that they sum up to $1$ million. The results are shown in Fig.~\ref{fig:toy_1D_multi} and corresponding Tab.~\ref{tab:toy_1D_multi}. It is quite clear that as the number of generators are increased up to $4$, the sampling keeps getting more realistic. In case when multiple modes are significantly overlapped/clustered, a generator can capture cluster of modes. Therefore, for this real data distribution, $4$ generators are enough to capture all the $5$ modes. With $5$ or more generators, all the modes were still captured, but the two overlapping modes have more than two generation peaks. This is mainly because multiple generators are capturing this region and all the generators (mixture components) were assigned {\em equal weights} during sampling.

Other works using more than one generators~\cite{liu2016coupled, arora2017generalization} also use the number of generators as a hyper-parameter as knowing a-priori the number of modes in a real-world data (\eg images) in itself is an open problem.

	\begin{table}
		\centering
		\begin{tabular}{c@{\hspace{2pt}}c@{\hspace{8pt}}c} 
			\toprule
			\textbf{GAN Variants} & \textbf{KL Div} & \textbf{\# Modes Covered} \\
			\midrule
			DCGAN~\cite{radford2015unsupervised}  & 2.15  & 712\\ 
			WGAN~\cite{arjovsky2017wasserstein}  & 1.02 & 868 \\ 
			BEGAN~\cite{berthelot2017began}  & 1.89 & 819\\ 
			GoGAN~\cite{juefei2017gang}  & 2.89 & 672\\
			Unrolled GAN~\cite{metz2017unrolledGAN} &1.29 & 842\\ 
			Mode-Reg DCGAN~\cite{che2016mode}  & 1.79 & 827 \\ 
			InfoGAN~\cite{chen2016infogan}  & 2.75 & 840 \\ 
			MA-GAN & 3.4 & 700 \\
			\textbf{MAD-GAN (Our)} &  \textbf{0.91} & \textbf{890}\\ 
			\bottomrule		
		\end{tabular}
		\caption{\label{tab:stacked_mnist}Stacked-MNIST experiments and comparisons. Note that three generators are used for MAD-GAN.
		}
		\vspace{-3mm}
	\end{table}

	\begin{table}
	\centering
		\begin{tabular}{c@{\hspace{2pt}}c@{\hspace{8pt}}c} 
			\toprule
			\textbf{GAN Variants} & \textbf{KL Div} & \textbf{\# Modes Covered} \\
			\midrule
			DCGAN~\cite{radford2015unsupervised}  & 0.18  & 980\\ 
			WGAN~\cite{arjovsky2017wasserstein} & 0.25 & 1000\\ 
			BEGAN~\cite{berthelot2017began} & 0.19 & 999\\ 
			GoGAN~\cite{juefei2017gang} & 0.87 & 972\\
			Unrolled GAN~\cite{metz2017unrolledGAN} &0.091 & 1000 \\
			Mode-Reg DCGAN~\cite{che2016mode}  & 0.12 & 992 \\ 
			InfoGAN~\cite{chen2016infogan}  & 0.47 & 990 \\ 
			MA-GAN & 1.62 & 997 \\
			\textbf{MAD-GAN (Our)} &  \textbf{0.074} & \textbf{1000}\\ 
			\bottomrule
		\end{tabular}
		\caption{\label{tab:compositional_mnist}Compositional-MNIST experiments and comparisons. Note that three generators are used for MAD-GAN.
		}
		\vspace{-3mm}
	\end{table}

\newcommand{\addSubFigHalf}[2]{\begin{subfigure}{.48\linewidth}
   \centering
   \includegraphics[width=.88\linewidth]{#1}
   \label{#2}\end{subfigure}
}

\begin{figure*}
\centering
    \addSubFigHalf{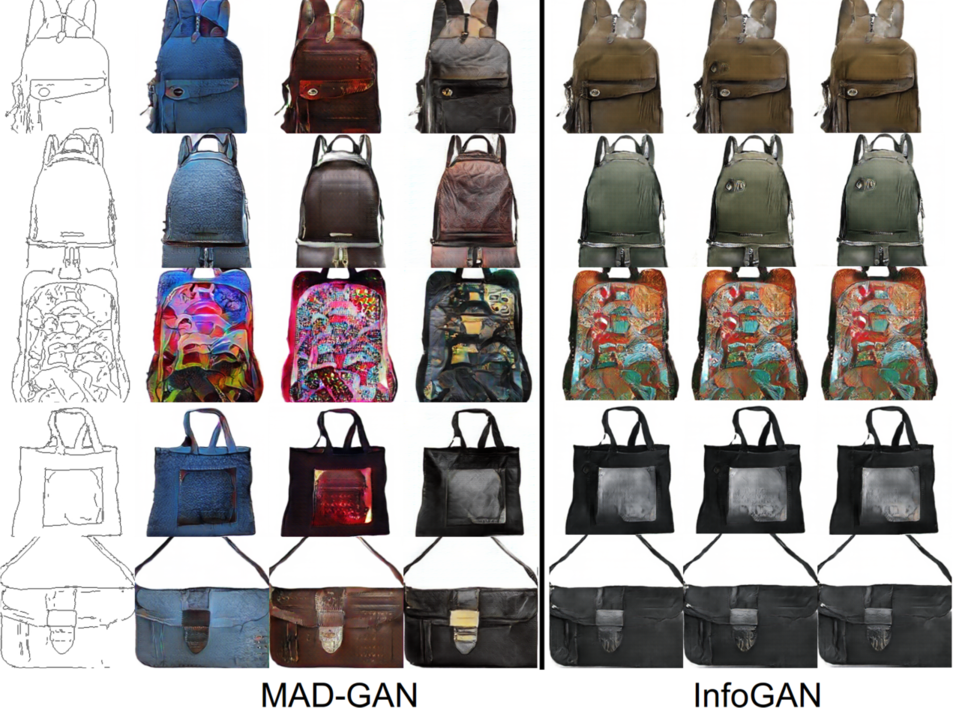}{fig:bags1} 
    \addSubFigHalf{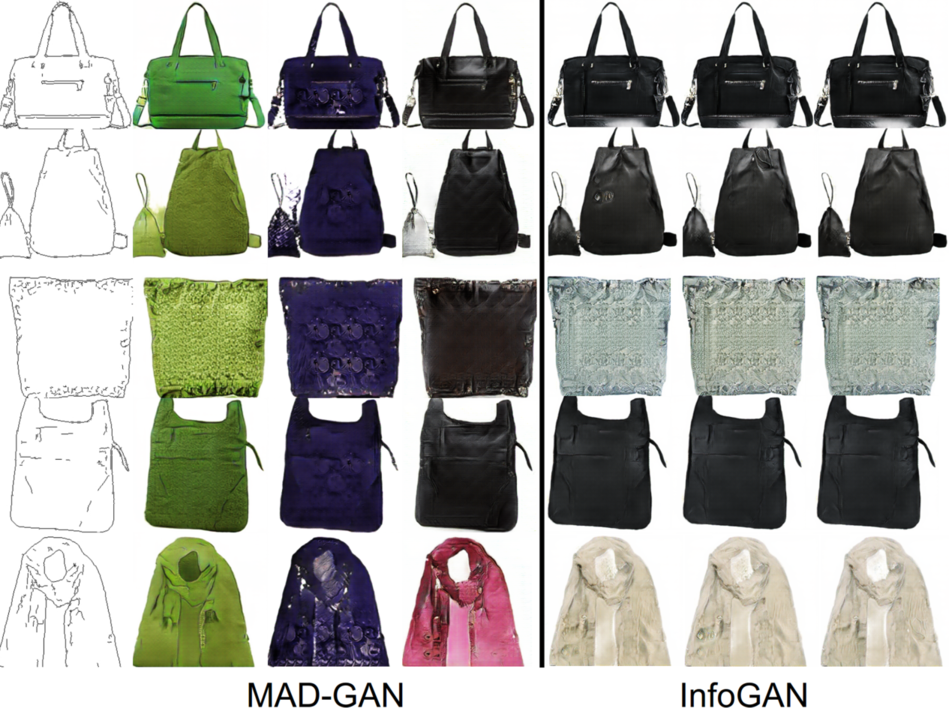}{fig:bags2} \\
    \captionof{figure}{Diverse generations for {\em edges-to-handbags} generation task. In each sub-figure, the first column is the input, columns 2-4 are generations by MAD-GAN (using three generators), and columns 5-7 are generations by InfoGAN (using three categorical codes). Clearly different generators of MAD-GAN are producing diverse results capturing different colors, textures, design patterns, \etc. However, InfoGAN generations are visually almost the same, indicating mode collapse.
    }\label{fig:pix2pixBag}
    \vspace{-2mm}
\end{figure*}

\begin{figure*}
\centering
    \addSubFigHalf{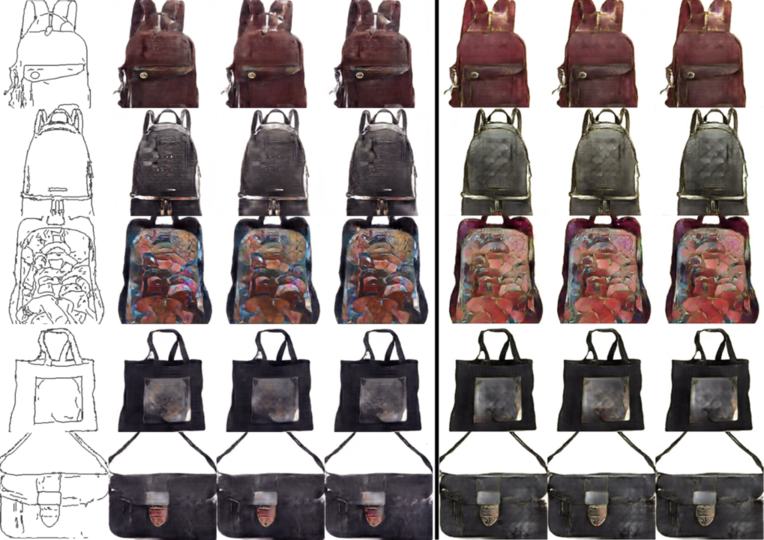}{fig:infobags1} 
    \addSubFigHalf{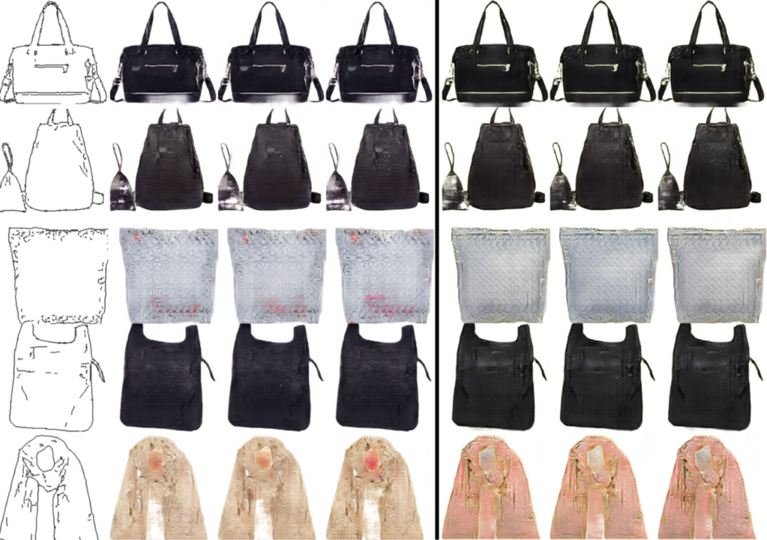}{fig:infobags2} \\
    \captionof{figure}{InfoGAN for {\em edges-to-handbags} task by sharing the discriminator and Q Network. In each sub-figure, the first column is the input, columns $2-4$ are generations when input is categorical code besides conditioning image, and columns $5-7$ are generations with noise as an additional input. The generations for both the architectures are visually the same irrespective of the categorical code value, which clearly indicates that it is not able to capture diverse modes.
    }\label{fig:pix2pixBag_infoGAN}
    \vspace{-3mm}
\end{figure*}

\subsection{Stacked and Compositional MNIST}
We now perform experiments on a more challenging setup, similar to~\cite{che2016mode,metz2017unrolledGAN}, in order to examine and compare MAD-GAN with other GAN variants. \cite{metz2017unrolledGAN} created a Stacked-MNIST dataset with $25,600$ samples where each sample has three channels stacked together with a random MNIST digit in each of them. Thus, it creates $1000$ distinct modes in the data distribution. \cite{metz2017unrolledGAN} used a stripped down version of the generator and discriminator pair to reduce the modeling capacity. We do the same for fair comparisons and used the same architecture as mentioned in their paper. Similarly, \cite{che2016mode} created Compositional-MNIST whereby they took $3$ random MNIST digits and place them at the $3$ quadrants of a $64 \times 64$ dimensional image. This also resulted in a data distribution with $1000$ hand-designed modes. The distribution of the resulting generated samples was estimated using a pretrained MNIST classifier to classify each of the digits either in the channels or the quadrants to decide the mode it belongs to. 

Tables~\ref{tab:stacked_mnist} and~\ref{tab:compositional_mnist} provide comparison of our method with variants of GAN in terms of KL divergence and the number of modes recovered for the Stacked and Compositional MNIST datasets, respectively. In Stacked-MNIST, as evident from the Tab.~\ref{tab:stacked_mnist}, MAD-GAN outperforms all other variants of GAN in both the criteria. Interestingly, in the case of Compositional-MNIST, as shown in Tab.~\ref{tab:compositional_mnist}, MAD-GAN, WGAN and Unrolled GAN were able to recover all the $1000$ modes. However, in terms of KL divergence, the distribution generated by MAD-GAN is the closest to the true data distribution.

\begin{figure}
    \centering
    \addSubFigHalf{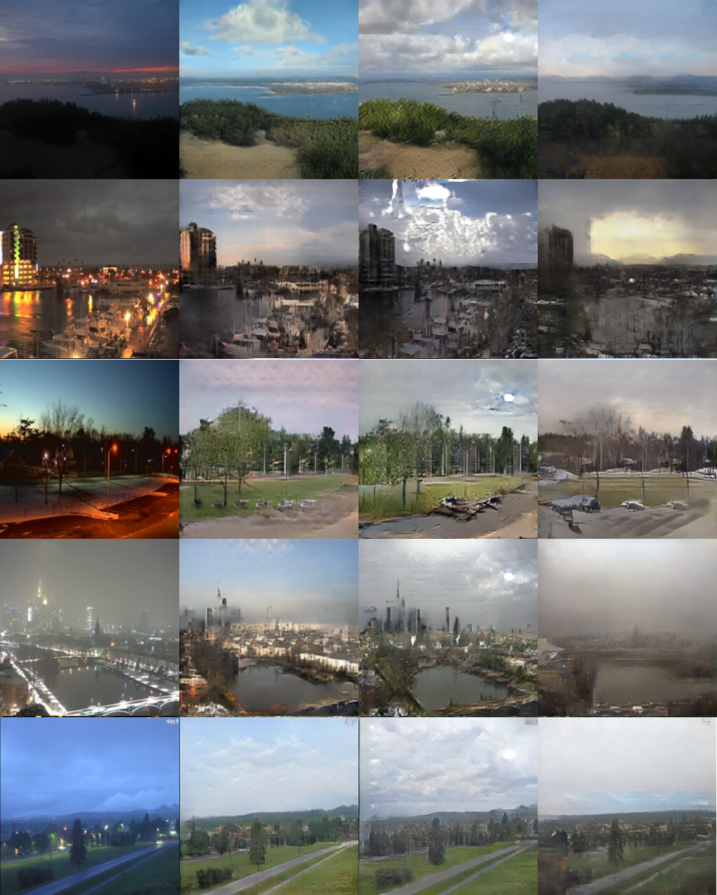}{fig:nightCom} 
    \addSubFigHalf{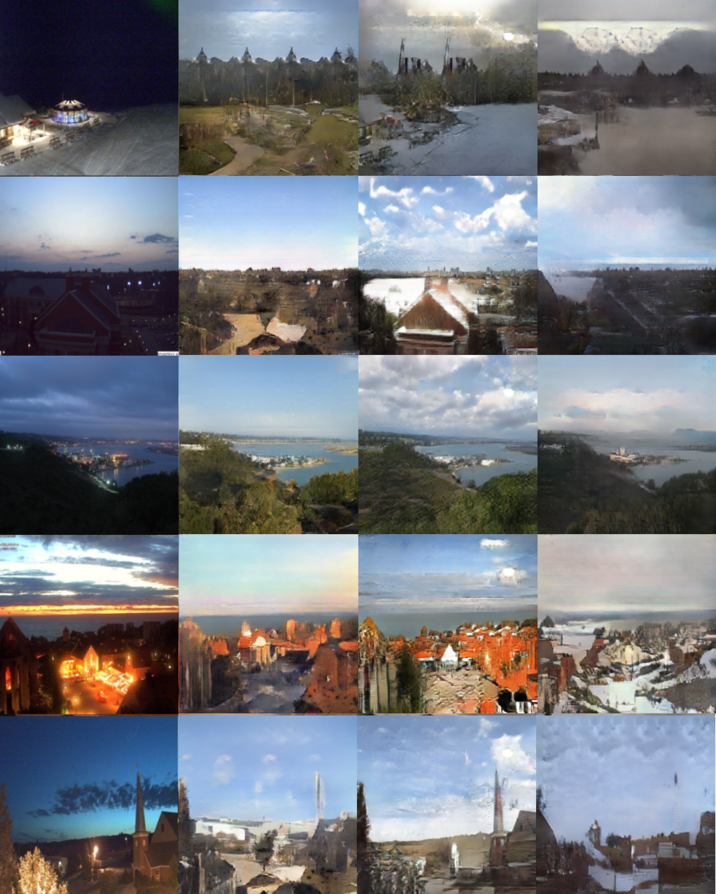}{fig:nightGen} 
    \captionof{figure}{Diverse generations for {\em night-to-day} image generation task. First column in each sub-figure represents the input. The remaining three columns show the diverse generations of three different generators of MAD-GAN (Our). 
    }\label{fig:image2imageNight}
    \vspace{-4mm}
\end{figure}

\newcommand{\addSubFigThird}[2]{\begin{subfigure}{.31\linewidth}
   \centering
   \includegraphics[width=\linewidth]{#1}
   \label{#2}\vspace{-3mm}\end{subfigure}
}

\begin{figure*}
\centering
    \addSubFigThird{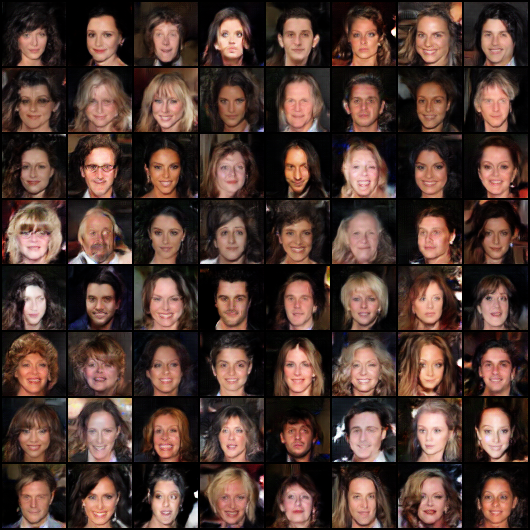}{fig:fg1} 
    \addSubFigThird{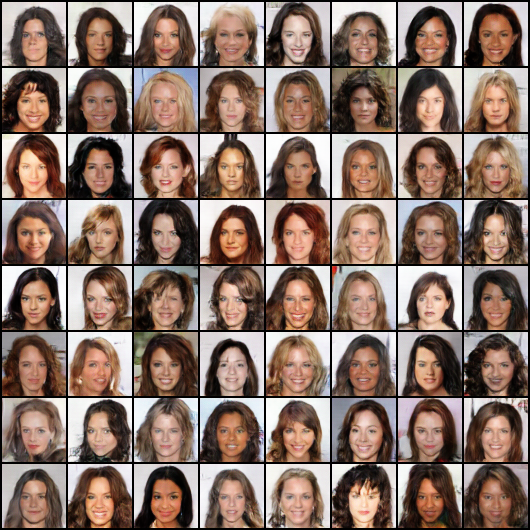}{fig:fg2}
    \addSubFigThird{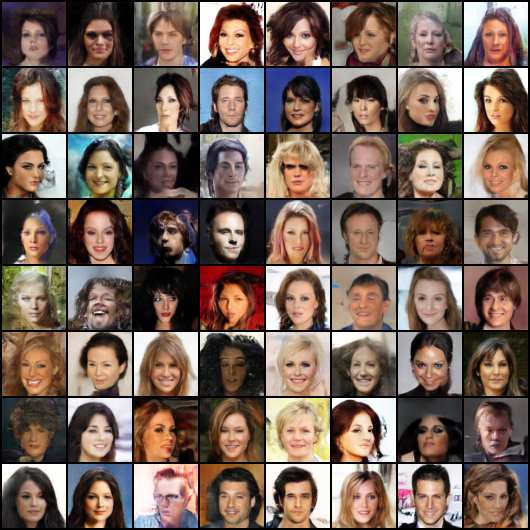}{fig:fg3}
    \captionof{figure}{Face generations using MAD-GAN. Each sub-figure shows generations by a single generator. The first generator is generating faces with very dark background. The second one is generating female faces with long hair in light background, while the third one is generating faces with colored background and casual look (based on facial direction and expression).
    }\label{fig:faces}
    \vspace{-3mm}
\end{figure*}

\subsection{Diverse Samples for Image-to-Image Translation and Comparison to InfoGAN}
Here we present experimental results on the challenging task of image-to-image translation~\cite{isola2016image2image} which uses conditional variant of GANs~\cite{mirza2014conditional}. Conditional GAN for this task is known to learn the delta distribution, thus, generates the same image irrespective of the variations in the input noise vector. Generating diverse samples in this setting in itself is an open problem. We show that MAD-GAN is able to generate diverse samples in these experiments as well. We use three generators for MAD-GAN experiments and show three diverse generations. Note that, we do not claim to capture all the possible modes present in the data distribution because firstly we cannot estimate the number of modes a priori, and secondly, even if we could, we do not know how diverse the generations would be after using certain number of generators. 
We follow the same approach as~\cite{isola2016image2image} and employ patch based conditional GAN. 

We compare MAD-GAN with InfoGAN~\cite{chen2016infogan} in these experiments as it is closest to our approach and can be used in image-to-image translation task. Theoretically, latent codes in InfoGAN should enable diverse generations. However, InfoGAN can only be used when the bias introduced by the categorical variables have significant impact on the generator network. For image-to-image translation and high resolution generations, the categorical variable does not have sufficient impact on the generations. As will be seen shortly, we validate this hypothesis by comparing our method with InfoGAN for this task. 
For the InfoGAN generator, to capture three kinds of distinct modes, the categorical code is chosen to take three values. Since we are dealing with images, in  this case, the categorical code is a 2D matrix in which we set one third of the entries to 1 and remaining to 0 for each category. The generator is fed input image along with categorical code appended channel wise to the image. Architecture of the Q network is same as that of the pix2pix discriminator~\cite{isola2016image2image}, except that the output is a vector of size $3$ for the prediction of the categorical codes. Note that, we tried different variations of the categorical codes but did not observe any significant variation in the generations.

Fig.~\ref{fig:pix2pixBag} shows generations by MAD-GAN and InfoGAN for the {\em edges-to-handbags} task, where given the edges of handbags, the objective is to generate real looking handbags. Clearly, each MAD-GAN generator is able to produce meaningful images but different from remaining generators in terms of color, texture, and patterns. However, InfoGAN generations are almost the same for all the three categorical codes. The results shown for InfoGAN are obtained by not sharing the discriminator and Q network parameters.

To make our baseline as strong as possible, we did some more experiments with InfoGAN for the {\em edges-to-handbags} task. For Fig.~\ref{fig:pix2pixBag_infoGAN}, we did two experiments by sharing all the initial layers of the discriminator and Q network. In the first experiment, the input is the categorical code besides the conditional image. In the second experiment, noise is also added as an input. The architecture details are given in Appendix~\ref{subsection:InfoArch}. In Fig.~\ref{fig:pix2pixBag_infoGAN}, we show the results of both these experiments side by side. There are not much perceivable changes as we vary the categorical code values. Generator simply learn to ignore the input noise as was also pointed by~\cite{isola2016image2image}.

In addition, in Fig.~\ref{fig:image2imageNight}, we show diverse generations for the {\em night-to-day} task, where given night images of places, the objective is to generate their corresponding day images. As can be seen, the generated day images in Fig.~\ref{fig:image2imageNight} differ in terms of lighting conditions, sky patterns, weather conditions, and many other minute yet useful cues.

\newcommand{\addSubFigDCGAN}[3]{\begin{subfigure}[t]{.7\linewidth}
   \includegraphics[width=\linewidth]{#1}
   \caption{#2}\label{#3}\end{subfigure}
}
\begin{figure}
    \centering
    \includegraphics[width=\linewidth]{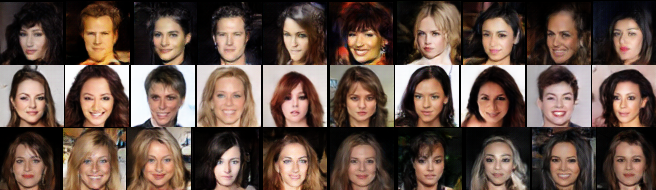}
    \caption{Face generations using MAD-GAN. Each generator employed is DCGAN. Each row represents a generator. Each column represents generations for a given random noise input $z$. Note that, the first generator is generating faces pointing to the left. The second generator is generating female faces with long hair, while the third generator generates images with light background. 
    }
        \label{fig:DCGANfaces}
        \vspace{-4mm}
\end{figure}

\subsection{Diverse-Class Data Generation}
To further explore the mode capturing capacity of MAD-GAN, we experimented with a much more challenging task of diverse-class data generation. In detail, we trained MAD-GAN (three generators) on a combined dataset consisting of various highly diverse images such as {\em islets}, {\em icebergs}, {\em broadleaf-forest}, {\em bamboo-forest}, and {\em bedroom}, obtained from the Places dataset~\cite{zhou2017places}. Images were randomly selected from each of them, creating a training dataset of $24,000$ images. The generators have the same architecture as that of DCGAN. In this case, as the images in the dataset belong to different classes, we did not share the generator parameters. As shown in Fig.~\ref{fig:madgan_multiview}, to our surprise, we found that even in this highly challenging setting, the generations from different generators belong to different classes. This clearly indicates that the generators in MAD-GAN are able to disentangle inter-class variations. In addition, each generator for different noise input is able to generate diverse samples, indicating intra-class diversity. 

\subsection{Diverse Face Generation}
Here we show diverse face generations (CelebA dataset) using MAD-GAN where we use DCGAN~\cite{radford2015unsupervised} as each of our three generators. Again, we use the same setting as provided in DCGAN. The high quality face generations are shown in the Fig.~\ref{fig:DCGANfaces}. 

To get better understanding about the possible diversities, we show additional generations in Fig.~\ref{fig:faces}.

\subsection{Unsupervised Representation Learning}
\label{subsec:unsup_svhn}
Similar to DCGAN~\cite{radford2015unsupervised}, we train our framework using SVHN dataset~\cite{Netzer11SVHN}. The trained discriminator is used to extract features. Using these features, we train an SVM for the classification task. For the MAD-GAN, with three generators, we obtained misclassification  error of $17.5 \%$ which is almost $5\%$ better than the results reported by DCGAN ($22.48 \%$). This clearly indicates that our framework is able to learn a better feature space in an unsupervised setting. 

\section{Conclusion}
We presented a very simple and effective framework, Multi-Agent Diverse GAN (MAD-GAN), for generating diverse and meaningful samples. We showed the efficacy of our approach and compared it with various variants of GAN that it captures diverse modes while producing high quality samples. We presented a theoretical analysis of MAD-GAN with conditions for global optimality. Looking forward, an interesting future direction would be to estimate a priori the number of generators needed for a particular dataset. It is not clear how to do that given that we do not have access to the true data distribution. In addition, we would also like to theoretically understand the limiting cases that depend on the relationship between the number of generators and the complexity of the data distribution. Another interesting direction would be to exploit different generators such that their combinations can be used to capture diverse modes. 

\section{Acknowledgements}
This work was supported by the EPSRC, ERC grant ERC-2012-AdG 321162-HELIOS, EPSRC grant Seebibyte EP/M013774/1 and EPSRC/MURI grant EP/N019474/1.

{\small
\bibliographystyle{ieee}
\bibliography{biblio}
}

\newpage
\appendix

\section*{Appendix}
Here, we first give better insights about the Theorem~\ref{theo:optimalConditionGenId} and discuss how and when MAD-GAN leads to diverse generations. In Appendix~\ref{sec:competing}, we introduce another way of getting different generators to generate diverse samples. We introduce intuitive similarity based competing objective (MAD-GAN-Sim) which encourages different generators to generate diverse samples. Finally in Appendix~\ref{sec:architectures}, we provide architecture details and data preparations for all the experiments reported for MAD-GAN and MAD-GAN-Sim.

\section{Insights for Diversity in MAD-GAN}
One obvious question that could arise is that {\em is it possible that all the generators learn to capture the same mode?}. The short answer is, theoretically {\em yes} and in practice {\em no}. Let us begin with the discussion to understand this. Theoretically, according to Theorem~\ref{theo:optimalConditionGenId} if $p_{g_i} = p_d$, for all $i$, then also the minimum objective value can be achieved. This implies, in worst case, MAD-GAN would perform same as the standard GAN. However, as discussed below, this is possible in following {\em highly unlikely} situations:
\begin{itemize}
    \item all the generators {\em always generate exactly similar} samples so that the discriminator is not able to differentiate them. In this case, the discriminator will learn a uniform distribution over the generator indices, thus, the gradients passed through the discriminator will be exactly the same for all the generators. However, this situation in general is not possible as all the generators are initialized differently. Even a slight variation in the samples from the generators will be enough for the discriminator to identify them and pass different gradient information to each generator. In addition, the objective function of generators is {\em only} to generate {\em real} samples, thus, there is nothing that encourages them to generate exactly the same samples. 
    \item the discriminator does not have enough capacity to learn the optimal parameters. This is in contrast to the assumption made in Theorem 1, which is that the discriminator is {\em optimal}. Thus, it should have enough capacity to learn a feature representation such that it can correctly identify samples from different generators. In practice, this is a very easy task and we did not have to modify anything up to the feature representation stage of the architecture of the discriminator. We used the standard architectures (explained in Appendix~\ref{sec:architectures}) for all the tasks. 
\end{itemize}
Hence, with random initializations and sufficient capacity generator/discriminator, we can easily avoid the trivial solution in which all the generators focus on exactly the same region of the true data distribution. This has been very clearly supported by various experiments showing diverse generations by MAD-GAN. 

\section{Similarity based competing objective}
\label{sec:competing}
We have discussed the MAD-GAN architecture using generator identification based objective. In this section, we propose a different extension to the standard GAN : {\em similarity based competing objective}, which we call as MAD-GAN-Sim. Here, we augment the GAN objective function with a diversity enforcing term. It ensures that the generations from different generators are diverse where the diversity depends on a user-defined task-specific function.

\begin{figure}[ht]
    \centering
     \includegraphics[scale=0.35]{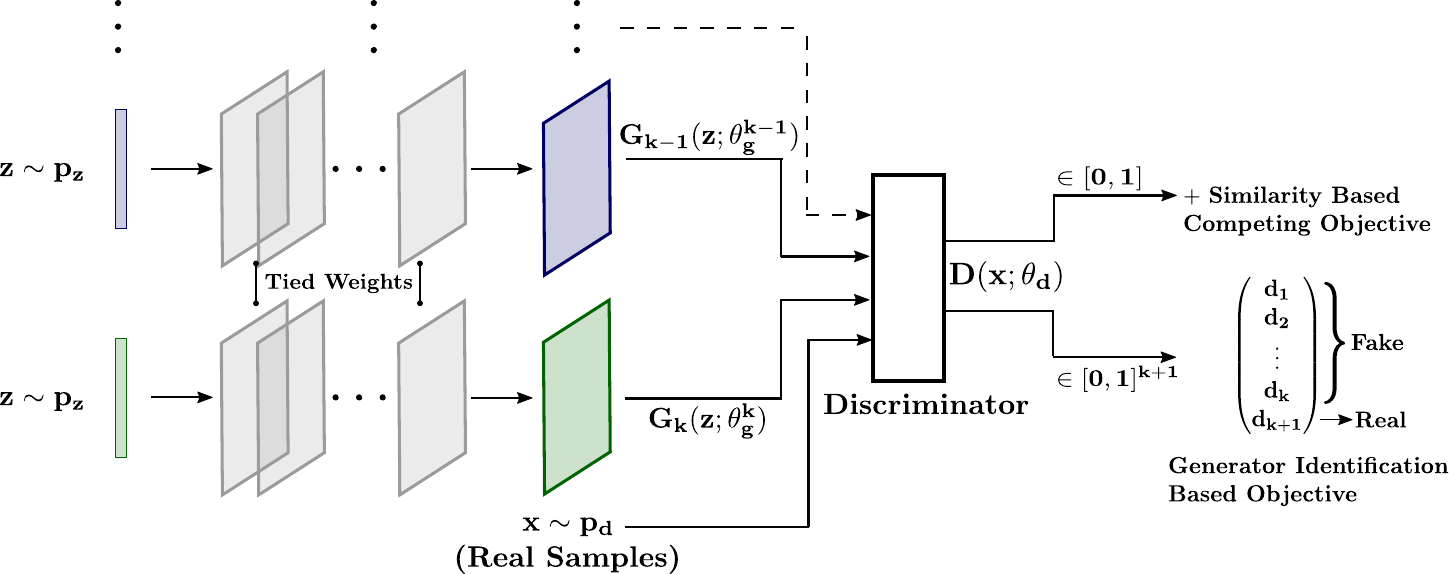}
	\caption{MAD-GAN-Sim compared with MAD-GAN. All the generators share parameters of all the layers except the last one. Two proposed diversity enforcing objectives, `competing' (MAD-GAN-Sim) and `generator identification' (MAD-GAN), are shown at the end of the discriminator}
   \label{fig:multiAgentGAN_both}
\end{figure}

The architecture is same as MAD-GAN discussed in Section~\ref{subsection:architecture} (refer Fig.~\ref{fig:multiAgentGAN_both}).

\subsection{Approach}
The approach presented here is motivated by the fact that the samples from different modes must look different. For example, in the case of images, these samples should differ in terms of texture, color, shading, and various other cues. Thus, different generators must generate dissimilar samples where the dissimilarity comes from a task-specific function. Before delving into the details, let us first define some notations in order to avoid clutter. We denote $\theta_g^i$ as the parameters of the $i$-th generator. The set of generators is denoted as $K = \{1, \cdots, k\}$. Given random noise $z$ to the $i$-th generator, the corresponding generated sample $G_i(z; \theta_g^i)$ is denoted as $g_i(z)$. Using these notations and following the above discussed intuitions, we impose following constraints over the $i$-th generator while updating its parameters:
\begin{align}
    \label{eq:divConstraint}
&D( G_i( z; \theta_g^i ); \theta_d ) \geq D ( G_j (z; \theta_g^j ) ; \theta_d) \nonumber \\
&+ \Delta\big( \phi(g_i(z)), \phi(g_j(z)) \big), \; \; \forall j \in K \setminus i
\end{align}
where, $\phi(g_i(z))$ denotes the mapping of the generated image $g_i(z)$ by the $i$-th generator into a feature space and $\Delta(.,.) \in [0,1]$ is the similarity function. Higher the value of $\Delta(.,.)$ more similar the arguments are. Intuitively, the above set of constraints ensures that the discriminator score for each generator should be higher than all other generators with a margin proportional to the similarity score. If the samples are similar, the margin increases and the constraints become more active. We use unsupervised learning based representation as our mapping function $\phi(.)$. Precisely, given a generated sample $g_i(z)$, $\phi(g_i(z))$ is the feature vector obtained using the discriminator of our framework. This is motivated by the feature matching based approach to improve the stability of the training of GANs~\cite{salimans2016improved}. The $\Delta(.,.)$ function used in this work is the standard cosine similarity based function. The above mentioned constraints can be satisfied by maximizing an equivalent unconstrained objective function as defined below:
\begin{align*}
    &U(\theta_g^i, \theta_d) := f\Big( D( G_i( z; \theta_g^i ); \theta_d ) - \nonumber \\
   &\frac{1}{k-1} \sum_{j \in K \setminus i} \big( D( G_j (z; \theta_g^j ) ; \theta_d) + \Delta( \psi_i, \psi_j) \Big) 
\end{align*}
where, $f(a) = \min(0, a)$, $\psi_i = \phi(g_i(z))$, and  $\psi_j = \phi(g_j(z))$. Intuitively, if the argument of $f(.)$ is positive, then the desirable constraint is satisfied and there is no need to do anything. Otherwise, maximize the argument with respect to $\theta_g^i$. Note that instead of using all the constraints independently, we use the average of all of them. Another approach would be to use the constraint corresponding to the $j$-th generator that maximally violates the set of constraints shown in Eq.~\ref{eq:divConstraint}. Experimentally we found that the training process of the average constraint based objective is more stable than the maximum violated constraint based objective. The intuition behind using these constraints comes from the well know $1$-slack formulation of the structured SVM framework~\cite{Joachims09_CuttingPlane,Tsochantaridis04SSVM}. Thus, the overall objective for the $i$-th generator is:
\begin{align*}
    \min_{\theta_g^i} \; \; V(\theta_d, \theta_g^i) - \lambda \; U(\theta_g^i, \theta_d)
\end{align*}
where $\lambda \geq 0$ is the hyperparameter. Algorithm~\ref{algo:competingObjective} shows how to compute gradients corresponding to different generators for the above mentioned objective function. Notice that, once sampled, the same $z$ is passed through all the generators in order to enforce constraints over a {\em particular generator} (as shown in Eq.~\ref{eq:divConstraint}). However, in order for constraints to not to contradict with each other while updating another generator, a different $z$ is sampled again from the $p_z$. The Algorithm~\ref{algo:competingObjective} is shown for the batch of size one which can be trivially generalized for any given batch sizes. In the case of discriminator, the gradients will have exactly the same form as the standard GAN objective. The only difference is that in this case the fake samples are being generated by $k$ generators, instead of one.

\begin{algorithm}[tb]
\caption{Updating generators for MAD-GAN-Sim}
\label{algo:competingObjective}
\begin{algorithmic}[1]
\INPUT  $\theta_d$; $p(z)$; $\theta_g^i, \forall i \in \{1, \cdots, k \}$; $\lambda$.
\FOR {each generator $i \in \{1, \cdots, k\}$}
\STATE Sample noise from the given noise prior $z \sim p_z$.
\STATE Obtain the generated sample $G_i(z; \theta_g^i)$ and corresponding feature vector $\psi_i = \phi(G_i(z; \theta_g^i))$.
\STATE $\nu \leftarrow 0$.
\FOR {each generator $j \in \{1, \cdots, k\}\setminus i$}
\STATE Compute feature vector $\psi_j = \phi(G_j(z; \theta_g^j))$.
\STATE $\nu \leftarrow \nu +  D( G_j (z; \theta_g^j ) ; \theta_d) + \Delta(\psi_i, \psi_j)$.
\ENDFOR
\STATE $\nu \leftarrow D(G_i(z; \theta_g^i); \theta_d) - \frac{\nu}{k-1}$.
\IF {$\nu \geq 0$}
\STATE $\nabla_{\theta_g^i} \log (1 - D(G_i(z; \theta_g^i); \theta_d)))$.
\ELSE
\STATE $\nabla_{\theta_g^i} \big( \log (1 - D(G_i(z; \theta_g^i); \theta_d))) - \lambda U(\theta_g^i, \theta_d) \big)$ .
\ENDIF
\ENDFOR
\OUTPUT 
\end{algorithmic}
\end{algorithm}

\subsection{Experiments}
We present the efficacy of MAD-GAN-Sim on the real world datasets. 

\subsection{Diverse Samples for Image-to-Image Translation}
\begin{figure}
    \centering
    \addSubFigHalf{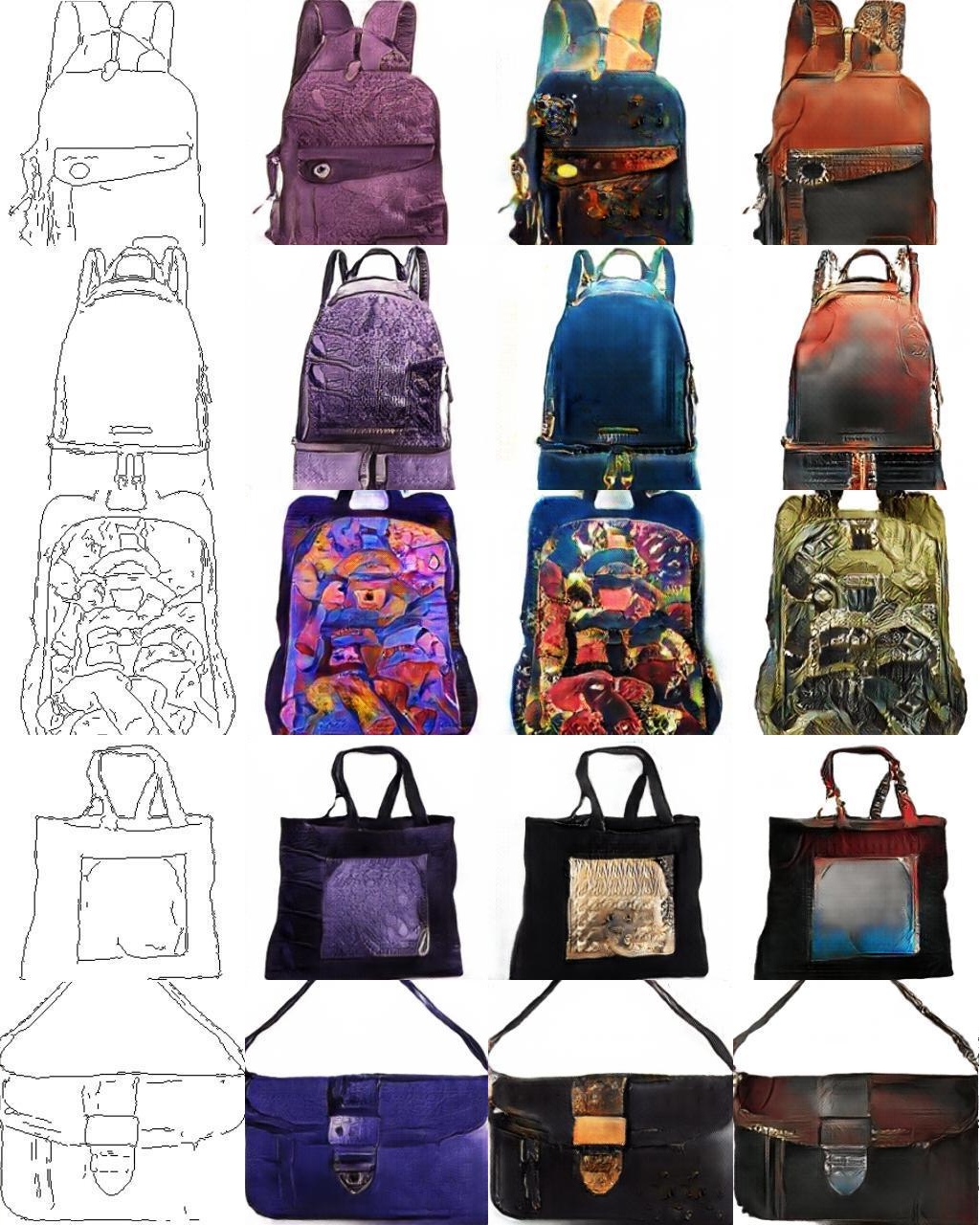}{fig:pix2pix_bags1_com} 
    \addSubFigHalf{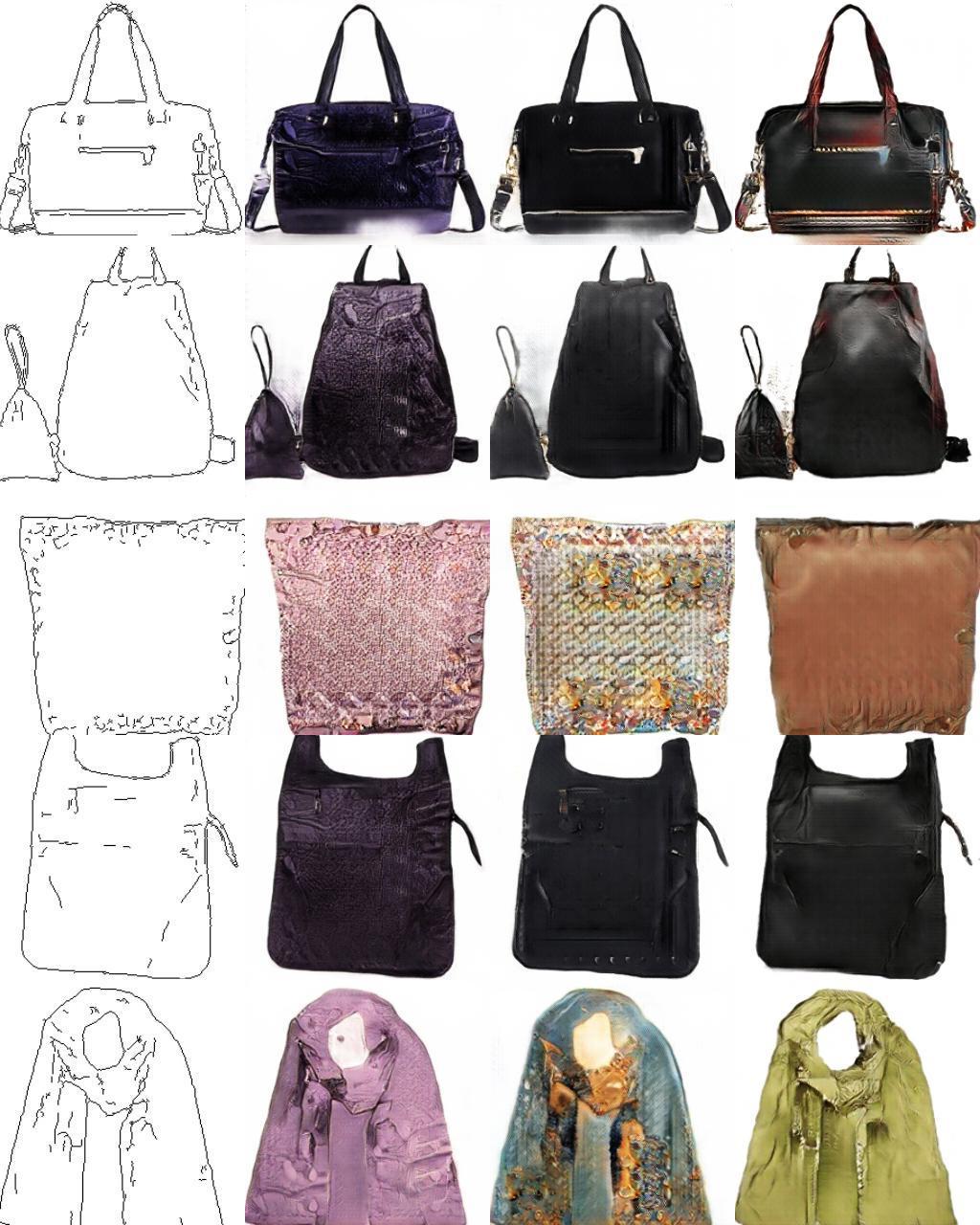}{fig:pix2pix_bags2_com} 
    \captionof{figure}{MAD-GAN-Sim: Diverse generations for `edges-to-handbags' image generation task. First column in each sub-figure represents the input. The remaining three columns show the diverse outputs of different generators. It is evident that different generators are able to produce very diverse results capturing color (brown, pink, black), texture, design pattern, shininess, among others. 
    }\label{fig:pix2pixBag_com}
\end{figure}

\begin{figure}
    \centering
    \addSubFigHalf{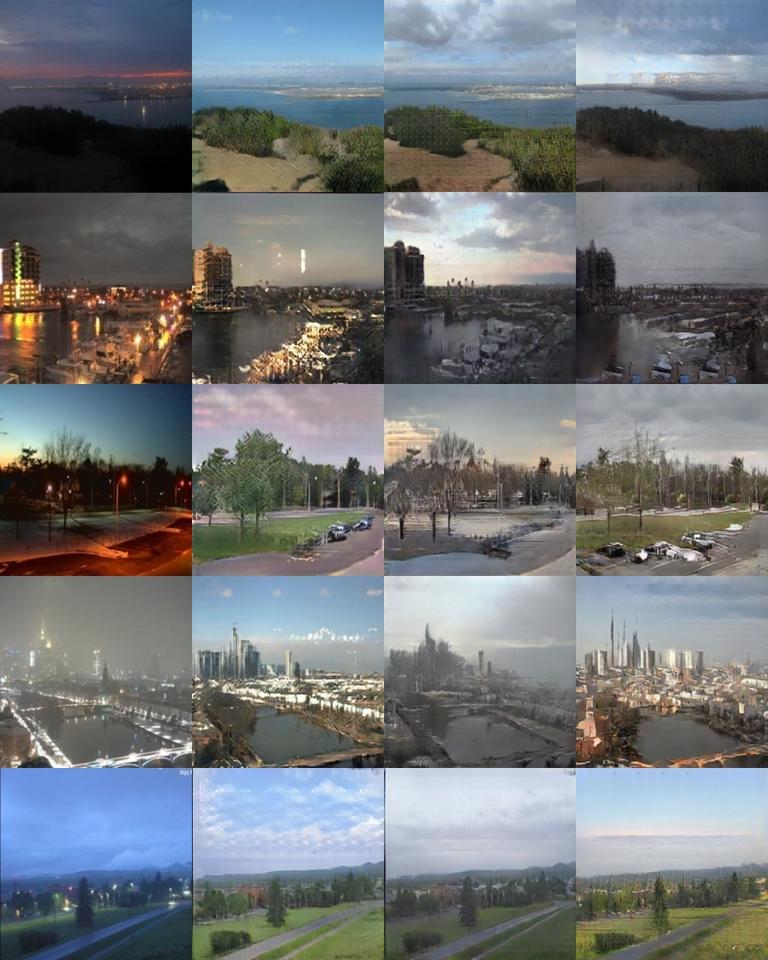}{fig:night2day_1_com} 
    \addSubFigHalf{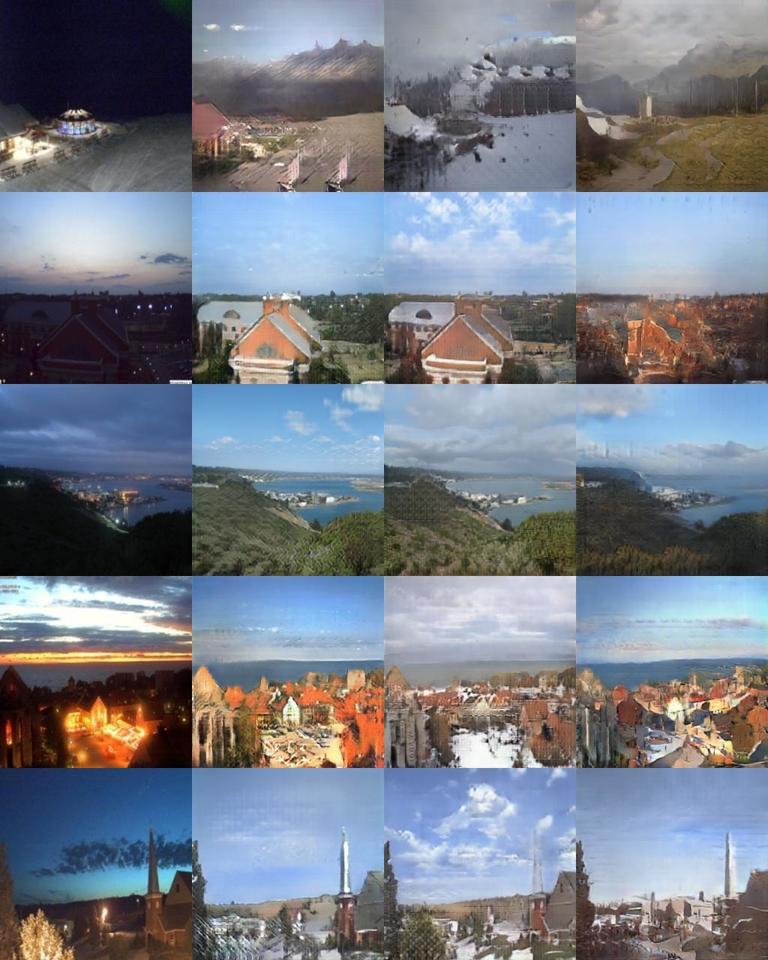}{fig:night2day_2_com} 
    \captionof{figure}{MAD-GAN-Sim: Diverse generations for `night to day' image generation task. First column in each sub-figure represents the input. The remaining three columns show the diverse outputs of different generators. It is evident that different generators are able to produce very diverse results capturing different lighting conditions, different sky patterns (cloudy vs clear sky), different weather conditions (winter, summer, rain), different landscapes, among many other minute yet useful cues. 
    }\label{fig:image2imageNight_com}
    \vspace{-2mm}
\end{figure}

We show diverse and highly appealing results using the diversity promoting objective. We use cosine based similarity to enforce diverse generations, an important criteria for real images. As before, we show results for the following two situations where diverse solution is useful: (1) given the edges of handbags, generate real looking handbags as in Fig.~\ref{fig:pix2pixBag_com}; and (2) given night images of places, generate their equivalent day images as in Fig.~\ref{fig:image2imageNight_com}. We clearly notice that each generator is able to produce meaningful and diverse images.

\subsection{Unsupervised Representation Learning}
We do the same experiment using SVHN as done in Section \ref{subsec:unsup_svhn}. For MAD-GAN-Sim we obtained the misclassification error of 18.3\% which is better than DCGAN (22.48\%). It clearly indicates that MAD-GAN-Sim is able to learn better feature representation in an unsupervised setting.

\section{Network Architectures and Parameters}
\label{sec:architectures}
Here we provide all the details about the architectures and the parameters used in various experiments. For the experiment concerning non-parametric density estimation, the MAD-GAN parameters are randomly initialized using xavier initialization with normal distributed random sampling \cite{glorot2010understanding}. For all the other experiments, the initialization done is same as the base architecture used to adapt MAD-GAN.

\begin{center}
\begin{table*}
\begin{tabular}{| m{10em} | m{5em} | m{5em} | m{4em} | m{4em} | m{4em} | m{4em} | m{4em} |}
 \hline
 \textbf{DCGAN, Unrolled GAN, InfoGAN, MA-GAN Disc} & \textbf{Mode-Reg DCGAN Disc} & \textbf{Mode-Reg DCGAN Enc} & \textbf{WGAN, GoGAN Disc} & \textbf{BEGAN Enc} & \textbf{BEGAN Dec} & \textbf{MAD-GAN Disc} & \textbf{InfoGAN QNet} \\
 \hline
 Input: 1 & 1 & 1 & 32 & 1 & 1 & 1 & 1 \\ 
 \hline
 \multicolumn{8}{|c|}{fc: 128, leaky relu}  \\
 \hline
 \multicolumn{8}{|c|}{fc: 128, leaky relu}  \\
 \hline
 fc: 1 & 1 & 64 & 1 & 32 & 1 & 5 (nGen+1) & 5 \\
 \hline
 \multicolumn{2}{|c|}{sigmoid} & \multicolumn{4}{|c|}{identity} & \multicolumn{2}{|c|}{softmax} \\
 \hline
\end{tabular}
\caption{\label{tab:toyDisc}Non-Parametric density estimation architecture for discriminators (Disc), encoders (Enc), decoders (Dec), and Q Network (QNet). nGen is number of generators, fc is fully connected layer.}
\end{table*}
\end{center}

\subsection{Non-Parametric Density Estimation}
\label{subsection:toy}
\paragraph{Architecture Details:}
The generator has two fully connected hidden layers with $128$ neurons (each of which are followed by exponential linear unit) and fully connected outer layer. In case of MAD-GAN and MA-GAN, we used $4$ generators with parameters of first two layers shared. Generator generates 1D samples. Input to each generator is a uniform noise $U(-1,1)$ of $64$ dimension. In case of InfoGAN, $5$ dimensional categorical code is further concatenated with the uniform noise to form the input. The categorical code is randomly sampled from the multinomial distribution. The discriminator architecture for respective networks is shown in Tab.~\ref{tab:toyDisc}. Mode-Regularized GAN architecture has encoder, BEGAN has encoder and decoder, and InfoGAN has Q Network whose details are also present in Tab.~\ref{tab:toyDisc}.

MAD-GAN has multi-label cross entropy loss. MA-GAN has binary cross entropy loss. For training, we use Adam optimizer with batch size of $128$ and learning rate of $1e-4$. In each mini batch, for MAD-GAN we have $128$ samples from each of the generators as well as real distribution, while for MA-GAN $128$ samples are chosen from real distribution as well as all the generators combined.

\paragraph{Dataset Generation}
We generated synthetic 1D data using GMM with $5$ Gaussians and select their means at $10, 20, 60, 80$ and $110$. The standard deviation used is $3,3,2,2$ and $1$. The first two modes overlap significantly while the fifth one is peaky and stands isolated.

\subsection{Stacked and compositional MNIST Experiments}
\label{subsection:stackMNIST}
\paragraph{Architecture details:} The architecture for stacked-MNIST is similar to the one used in \cite{metz2017unrolledGAN}. Please refer to the Tab.~\ref{tab:stackMNISTGen} for generator architecture and Tab.~\ref{tab:stackMNISTDisc} for discriminator architecture and Q network architecture of InfoGAN. The architecture for compositional-MNIST experiment is same as DCGAN \cite{radford2015unsupervised}. Please refer to the Tab.~\ref{tab:compositionalMNISTDisc} for discriminator architecture and Q network architecture of InfoGAN. In both the experiments, Q network of InfoGAN shares all except the last layer with the discriminator.
\begin{center}
\begin{table}
\begin{tabular}{| m{12em} | m{4em} | m{4em} |} 
 \hline
 & \textbf{number outputs} & \textbf{stride} \\
 \hline
 Input: $z \sim \mathcal{N}(0, I_{256})$ & & \\ 
 \hline
 Fully connected & 4 * 4 * 64 & \\ 
 \hline
 Reshape to image 4,4,64 & & \\
 \hline
 Transposed Convolution & 32 & 2 \\
 \hline
 Transposed Convolution & 16 & 2 \\
 \hline
 Transposed Convolution & 8 & 2 \\
 \hline
 Convolution & 3 & 1 \\
 \hline
\end{tabular}
\caption{\label{tab:stackMNISTGen}Generator architecture for $1000$ class stacked-MNIST experiment. For MAD-GAN, all the layers except those mentioned in last two rows are shared.}
\end{table}
\end{center}

\begin{center}
\begin{table}
\begin{tabular}{| m{12em} | m{4em} | m{4em} | } 
 \hline
 & \textbf{number outputs} & \textbf{stride} \\
 \hline
 Input: 32x32 Color Image  & & \\ 
 \hline
 Convolution & 4 & 2 \\ 
 \hline
 Convolution & 8 & 2 \\
 \hline
 Convolution & 16 & 2 \\
 \hline
 Flatten & & \\
 \hline
 Fully Connected & 1 & \\
 \hline
\end{tabular}
\caption{\label{tab:stackMNISTDisc} Discriminator architecture for $1000$ class stacked-MNIST experiment. For MAD-GAN, with $k$ generators, it is adapted to have $k+1$ dimensional last layer output. For InfoGAN, with 156 dimensional salient variables and 100 dimensional incompressible noise, it is adapted to have $156$ dimensional output for Q network. 
}
\end{table}
\end{center}

\begin{center}
\begin{table}
\begin{tabular}{| m{12em} | m{5em} | m{4em} | } 
 \hline
 & \textbf{number outputs} & \textbf{stride} \\
 \hline
 Input: Color Image (64x64) & & \\ 
 \hline
 Convolution & 64 & 2 \\ 
 \hline
 Convolution & 128 & 2 \\
 \hline
 Convolution & 256 & 2 \\
 \hline
 Convolution & 512 & 2 \\
 \hline
 Flatten & & \\
 \hline
 Fully Connected & 1 & \\
 \hline
\end{tabular}
\caption{\label{tab:compositionalMNISTDisc} Discriminator architecture for $1000$ class compositional-MNIST experiment. For MAD-GAN, with $k$ generators, it is adapted to have $k+1$ dimensional last layer output. For InfoGAN, with 156 dimensional salient variables and 100 dimensional incompressible noise, it is adapted to have $156$ dimensional output for Q network.
}
\end{table}
\end{center}

\paragraph{Dataset preparation:} MNIST database of hand written digits are used for both the tasks.

\subsection{Image-to-Image Translation}
\subsubsection{MAD-GAN/MAD-GAN-Sim}
\paragraph{Architecture details:}
The network architecture is adapted from~\cite{isola2016image2image} and the experiments were conducted with the U-Net architecture and patch based discriminator.

In more detail, let Ck denote a Convolution-BatchNorm-ReLU layer with k filters and CDk represent a Convolution-BatchNorm-Dropout-ReLU layer with a dropout rate of $50$\%. All Convolutions are $4\times 4$ spatial filters with a stride of $2$. Convolutions in the encoder, and in the discriminator, downsample by a factor of $2$, whereas in the decoder they upsample by a factor of $2$.

\paragraph{Generator Architectures}
We used the U-Net generator based architecture from~\cite{isola2016image2image} as follows:

\begin{itemize}
    \item U-Net Encoder: C64-C128-C256-C512-C512-C512-C512-C512
    \item U-Net Decoder: CD512-CD1024-CD1024-C1024-C1024-C512-C256-C128. Note that, in case of MAD-GAN, the last layer does not share parameters with other generators.
\end{itemize}
After the last layer in the decoder, a convolution is applied to map to the number of output channels to $3$, followed by a tanh function. BatchNorm is not applied to the first C64 layer in the encoder. All ReLUs in the encoder are leaky, with a slope of $0.2$, while ReLUs in the decoder are not leaky. The U-Net architecture has skip-connections between each layer $i$ in the encoder and layer $n-i$ in the decoder, where $n$ is the total number of layers. The skip connections concatenate activations from layer $i$ to layer $n-i$. This changes the number of channels in the decoder.

\paragraph{Discriminator Architectures}
The patch based $70\times 70$ discriminator architecture was used in this case : C64-C128-C256-C512.

\paragraph{Diversity term}
\begin{itemize}
\item MAD-GAN: After the last layer, a convolution is applied to map the output layer to the dimension of $k+1$ (where $k$ is the number of generators in MAD-GAN) followed by the softmax layer for the normalization. 
\item MAD-GAN-Sim: After the last layer, a convolution is applied to map to a 1 dimensional output followed by a Sigmoid function. For the unsupervised feature representation $\phi(.)$, the feature activations from the penultimate layer C256 of the discriminator was used as the feature activations for the computation of the cosine similarity. 
\end{itemize}

For the training, we used Adam optimizer with learning rate of $2e-4$ (for both generators and discriminator), $\lambda_{L1} = 10$ (hyperparameter corresponding to the $L_1$ regularizer), $\lambda = 1e-3$ (corresponding to MAD-GAN-Sim), and batch size of $1$.

\subsubsection{InfoGAN}
\label{subsection:InfoArch}
The network architecture is adapted from~\cite{isola2016image2image} and the experiments were conducted with the U-Net architecture and patch based discriminator.

\paragraph{Generator Architectures}
The U-Net generator is exactly same as in~\cite{isola2016image2image} except that the number of input channels are increased from $3$ to $4$. For the experiment done for Fig.~\ref{fig:pix2pixBag_infoGAN}, to take noise as input, input channels are increased to $5$ (one extra input channel for noise).

\paragraph{Discriminator Architectures}
The discriminator is exactly same as in~\cite{isola2016image2image}: C64-C128-C256-C512

\paragraph{Q network Architectures}
The Q network architecture is C64-C128-C256-C512-Convolution3-Convolution3. Here first Convolution3 gives a output of $30\times 30$ patches with $3$ channels while second Convolution3 just gives $3$ dimensional output. All the layers except last two are shared with the discriminator to perform the experiments for Fig.~\ref{fig:pix2pixBag_infoGAN}.

\paragraph{Diversity term}
To capture three kinds of distinct modes, the categorical code can take three values. Hence, in  this case, the categorical code is a 2D matrix in which one third of entries are set to $1$ and remaining to $0$ for each category. The generator is fed input image along with categorical code appended channel wise to the image. For the experiment done for Fig.~\ref{fig:pix2pixBag_infoGAN}, to take noise as input, the generator input is further channel wise appended with a 2D matrix of normal noise.

For the training, we used Adam optimizer with learning rate of $2e-4$ (for both generator and discriminator), $\lambda_{L1} = 10$ (hyperparameter corresponding to the $L_1$ regularizer) and batch size of $1$. 

\paragraph{Dataset Preparation:}
\begin{itemize}
\item Edges-to-Handbags:  We used $137,000$ Amazon handbag images from \cite{zhu2016generative}. The random split into train and test was kept the same as done by \cite{zhu2016generative}.
\item Night-to-Day: We used $17,823$ training images extracted from 91 webcams. We thank Jun-Yan Zhu for providing the dataset. 
\end{itemize}

\subsection{Diverse-Class Data Generation}
\paragraph{Architecture details:} The network architecture is adapted from DCGAN~\cite{radford2015unsupervised}. Concretely, the discriminator architecture is described in \tabref{tab:res-disc} and the generator architecture in \tabref{tab:dcgan-gen}. We use three generators without sharing any parameter. The residual layers helped in improving the image quality since the data manifold was much more complicated and the discriminator needed more capacity to accommodate it.

\paragraph{Diversity terms} For the training, we used Adam optimizer with the learning rate of $2e-4$ (both generator and discriminator) and batch size of $64$.

\paragraph{Dataset preparation:} Training data is obtained by combining dataset consisting of various highly diverse images such as {\em islets}, {\em icebergs}, {\em broadleaf-forest}, {\em bamboo-forest} and {\em bedroom}, obtained from the Places dataset~\cite{zhou2017places}. To create the training data, images were randomly selected from each of them, creating a dataset consisting of $24,000$ images.

\subsection{Diverse Face Generations with DCGAN}
\paragraph{Architecture details:} The network architecture is adapted from DCGAN~\cite{radford2015unsupervised}. Concretely, the discriminator architecture is described in \tabref{tab:res-disc} and the generator architecture in \tabref{tab:dcgan-gen}. In this case all the parameters of the generators except the last layer were shared. The residual layers helped in improving the image quality since the data manifold and the manifolds of each of the generators was much more complicated and the discriminator needed more capacity to accommodate it.

\begin{center}
\begin{table}
\begin{tabular}{ | m{22em} |} 
 \hline
 \textbf{Discriminator D} \\
 \hline
 Input 64x64 Color Image \\
 \hline
 4x4 conv. 64 leakyRELU. stride 2. batchnorm \\
 \hline
 4x4 conv. 128 leakyRELU. stride 2. batchnorm \\
 \hline
4x4 conv. 256 leakyRELU. stride 2. batchnorm \\
 \hline 
4x4 conv. 512 leakyRELU. stride 2. batchnorm \\  
 \hline
4x4 conv. output leakyRELU. stride 1 \\  
 \hline
\end{tabular}
\caption{\label{tab:dcgan-disc}DCGAN Discriminator: It is adapted to have $k+1$ dimensional last layer output for MAD-GAN with $k$ generators. (normalizer is softmax).}
\end{table}
\end{center}

\begin{center}
\begin{table}
\begin{tabular}{ | m{22em} | } 
 \hline
 \textbf{Generator G}\\
 \hline
 Input $\in \R^{100}$\\
 \hline
 4x4 upconv. 512 RELU.batchnorm.shared\\
 \hline
 4x4 upconv. 256 RELU. stride 2.batchnorm.shared\\
 \hline
 4x4 upconv. 128 RELU. stride 2.batchnorm.shared\\
 \hline 
 4x4 upconv. 64 RELU. stride 2.batchnorm.shared\\
 \hline
 4x4 upconv. 3 tanh. stride 2\\
 \hline
\end{tabular}
\caption{\label{tab:dcgan-gen}DCGAN Generator: All the layers except the last one are shared among all the three generators.}
\end{table}
\end{center}

\begin{center}
\begin{table}
\begin{tabular}{ | m{22em} |} 
 \hline
 \textbf{Residual Discriminator D} \\
 \hline
 Input 64x64 Color Image \\
 \hline
 7x7 conv. 64 leakyRELU. stride 2. pad 1. batchnorm \\
 \hline
 3x3 conv. 64 leakyRELU. stride 2. pad 1. batchnorm \\
 \hline
 3x3 conv. 128 leakyRELU. stride 2.pad 1.  batchnorm \\
 \hline
 3x3 conv. 256 leakyRELU. stride 2. pad 1. batchnorm \\
 \hline 
 3x3 conv. 512 leakyRELU. stride 2. pad 1. batchnorm \\  
 \hline
 3x3 conv. 512 leakyRELU. stride 2. pad 1. batchnorm \\ 
 \hline
 3x3 conv. 512 leakyRELU. stride 2. pad 1. batchnorm \\
 \hline
 RESIDUAL-(N512, K3, S1, P1) \\
 \hline
 RESIDUAL-(N512, K3, S1, P1) \\
 \hline
 RESIDUAL-(N512, K3, S1, P1) \\
 
 \hline
\end{tabular}
\caption{\label{tab:res-disc}Discriminator architecture for diverse-class data generation and diverse face generation: The last layer output is $k+1$ dimensional for MAD-GAN with $k$ generators (normalizer is softmax). `RESIDUAL' layer is elaborated in \tabref{tab:res-layer}.}
\end{table}
\end{center}

\begin{center}
\begin{table}
\begin{tabular}{ | m{22em} |} 
 \hline
 \textbf{RESIDUAL}-Residual Layer \\
 \hline
 \textbf{Input:} previous-layer-output \\
 \hline
 \textbf{c1:} CONV-(N512, K3, S1, P1), BN, ReLU \\
 \hline
 \textbf{c2:} CONV-(N512, K3, S2, P1), BN \\
 \hline
 SUM(c2,previous-layer-output) \\
 \hline
\end{tabular}
\caption{\label{tab:res-layer}Residual layer description for Tab.~\ref{tab:res-disc}.}
\end{table}
\end{center}

\paragraph{Diversity terms}

For the training, we used Adam optimizer with the learning rate of $2e-4$ (both generator and discriminator) and batch size of $64$.

\paragraph{Dataset preparation:} We used CelebA dataset as mentioned for face generation based experiments. For Image generation all the images ($14,197,122$) from the Imagenet-1k dataset~\cite{Deng09ImageNet} were used to train the DCGAN with $3$ Generators alongside the MAD-GAN objective. The images from both CelebA and Imagenet-1k were resized into $64\times 64$.

\subsection{Unsupervised Representation Learning}
\paragraph{Architecture details:} Our architecture uses the one proposed in DCGAN~\cite{radford2015unsupervised}. Similar to the DCGAN experiment on SVHN dataset ($32\times 32\times 3$)~\cite{Netzer11SVHN}, we removed the penultimate layer of generator (second last row in Tab.~\ref{tab:dcgan-gen}) and first layer of discriminator (first convolution layer in Tab.~\ref{tab:dcgan-disc}).

\begin{center}
\begin{table}
\begin{tabular}{ |m{4.5em} |m{4.5em} |m{4.5em} |m{4.5em} |} 
 \hline
 \textbf{Technique} & \textbf{2 Gen} & \textbf{3 Gen} & \textbf{4 Gen} \\
 \hline
 MAD-GAN & 20.5\% & 18.2\% & 17.5\% \\
 \hline
 MAD-GAN-Sim & 20.2\% & 19.6\% & 18.3\% \\
 \hline
\end{tabular}
\caption{\label{tab:dcgan-svhn} The misclassification error of MAD-GAN and MAD-GAN-Sim on SVHN while varying the number of generators.}
\end{table}
\end{center}

\paragraph{Classification task:} We trained our model on the available SVHN dataset \cite{Netzer11SVHN}. For feature extraction using discriminator, we followed the same method as mentioned in the DCGAN paper \cite{radford2015unsupervised}. The features were then used for training a regularized linear L2-SVM. The ablation study is presented in Tab.~\ref{tab:dcgan-svhn}.

\paragraph{Dataset preparation:} We used SVHN dataset \cite{Netzer11SVHN} consisting of $73,257$ digits for the training, $26,032$ digits for the testing, and $53,1131$ extra training samples. As done in DCGAN~\cite{radford2015unsupervised}, we used $1000$ uniformly class distributed random samples for training, $10,000$ samples from the non-extra set for validation and $1000$ samples for testing.

For the training, we used Adam optimizer with learning rate of $2e-4$ (both generator and discriminator), $\lambda = 1e-4$ (competing objective), and batch size of $64$.

\end{document}